\documentclass{article}
\usepackage{amssymb}
\usepackage{amsmath}
\usepackage{enumitem}
\usepackage{dsfont}
\usepackage{graphicx}
\usepackage{algorithm}
\usepackage{algpseudocode} 
\usepackage{subcaption} 
\usepackage{booktabs}
\usepackage{booktabs}
\usepackage{tabularx}
\usepackage{array}
\usepackage{placeins}

\usepackage[final]{corl_2026} 

\hypersetup{
  pdftitle={Simultaneous Forward and Inverse Human-in-the-Loop Optimization},
  pdfauthor={Kyeongwon Park and Steven H. Collins}
}

\title{Simultaneous Forward and Inverse Human-in-the-Loop Optimization}

\author{
  Kyeongwon~Park\\
  Department of Mechanical Engineering\\
  Stanford University, 
  United States\\
  \texttt{parkkw@stanford.edu} \\
  \And
  Steven H.~Collins\\
  Department of Mechanical Engineering\\
  Stanford University, 
  United States\\
  \texttt{stevecollins@stanford.edu} \\
}

\begin{document}
\maketitle


\begin{abstract}
Subjective user experience is important to human--robot interaction, but the outcomes users value, and how those preferences vary across individuals and contexts, are often unknown. While inverse learning approaches using human data can help identify user rewards, in many assistive settings the experimental costs of executing a control policy, measuring biomechanical or physiological outcomes, and collecting user feedback often limit the number of queries and optimization iterations. Here, we present \textit{Simultaneous Forward and Inverse Human-In-the-Loop Optimization} (SFIHILO), which efficiently infers individual-specific reward functions from preferences over human outcomes and identifies a final control policy that maximizes the learned reward. SFIHILO bootstraps a forward model to predict user outcomes from control policies, uses this model for active querying to accelerate inverse reward learning, and then optimizes the final control policy without additional user trials. We score candidate policies by their expected reduction in uncertainty across both forward and inverse beliefs, targeting a regional preference boundary to robustly inform this simultaneous learning process. In simulation, we show that SFIHILO was effective across user heterogeneity, outcome dimensionalities, precision requirements, noise levels, and nonstationarity; compared with mutual information approaches, the proposed active querying strategy significantly improved sample efficiency in inverse learning while preserving forward model accuracy. This approach demonstrates the potential to infer latent human goals, enabling more transferable and effective human--robot interaction.

\end{abstract}

\section{Introduction}
What objectives, or rewards, should a robot pursue in human-in-the-loop contexts? When a robot's actions directly affect humans, its fundamental goal should be to prioritize human experience and outcomes rather than solely optimize internal device- or control-level metrics~\cite{tang2025deep, luo2024survey}. However, the outcomes people truly care about, and how those priorities vary across individuals, tasks, and contexts, remain largely unknown~\cite{young2016state, slade2024human}.

Inferring human rewards defined over human outcomes may substantially change how we align robot behavior. For example, prior robot personalization work has applied optimization frameworks that optimize human objective functions, such as energetic cost during daily tasks~\cite{zhang2017human, ding2018human, kim2017human, slade2022personalizing}, but the lack of knowledge about what individual users truly want to pursue in a given context limits their extension to real-world scenarios~\cite{lakmazaheri2024optimizing}. Preference-based human-in-the-loop optimization partially handles this by avoiding reliance on hand-crafted objectives and instead using pairwise comparisons to personalize robot behavior~\cite{lee2023user, thatte2017sample, arens2025preference, tucker2020preference}. However, because those methods typically treat the optimization process as a black-box function, we do not know \textit{why} users preferred certain control policies, so changes in devices or control settings typically require new user trials. Identifying individual human rewards can inform human-in-the-loop optimization for human--robot alignment, support easier transfer across settings, and, more fundamentally, help us better understand individual user goals and guide how we should design and train our robot systems.

Inverse learning approaches can help infer human rewards by identifying how reward features explain observed human data~\cite{ng2000algorithms, ziebart2008maximum, christiano2017deep, biyik2022aprel}. Standard reward-learning formulations, however, often assume either known dynamics~\cite{kim2021reward, brown2019extrapolating} or options that can already be described by a common set of features~\cite{biyik2022aprel, biyik2024active, katz2021preference}, assumptions that are difficult to satisfy when humans are part of the system. In such settings, each user's biomechanical and physiological characteristics make the mapping from control policies to reward features unknown, requiring physical execution of each control policy to measure the resulting outcomes. Because this lack of predictive capability also limits the a-priori evaluation of query informativeness, inverse learning for human rewards can impose a large experimental burden on users in practical applications.

\begin{figure*}[t]
    \centering
    \includegraphics[width=0.57\textwidth]{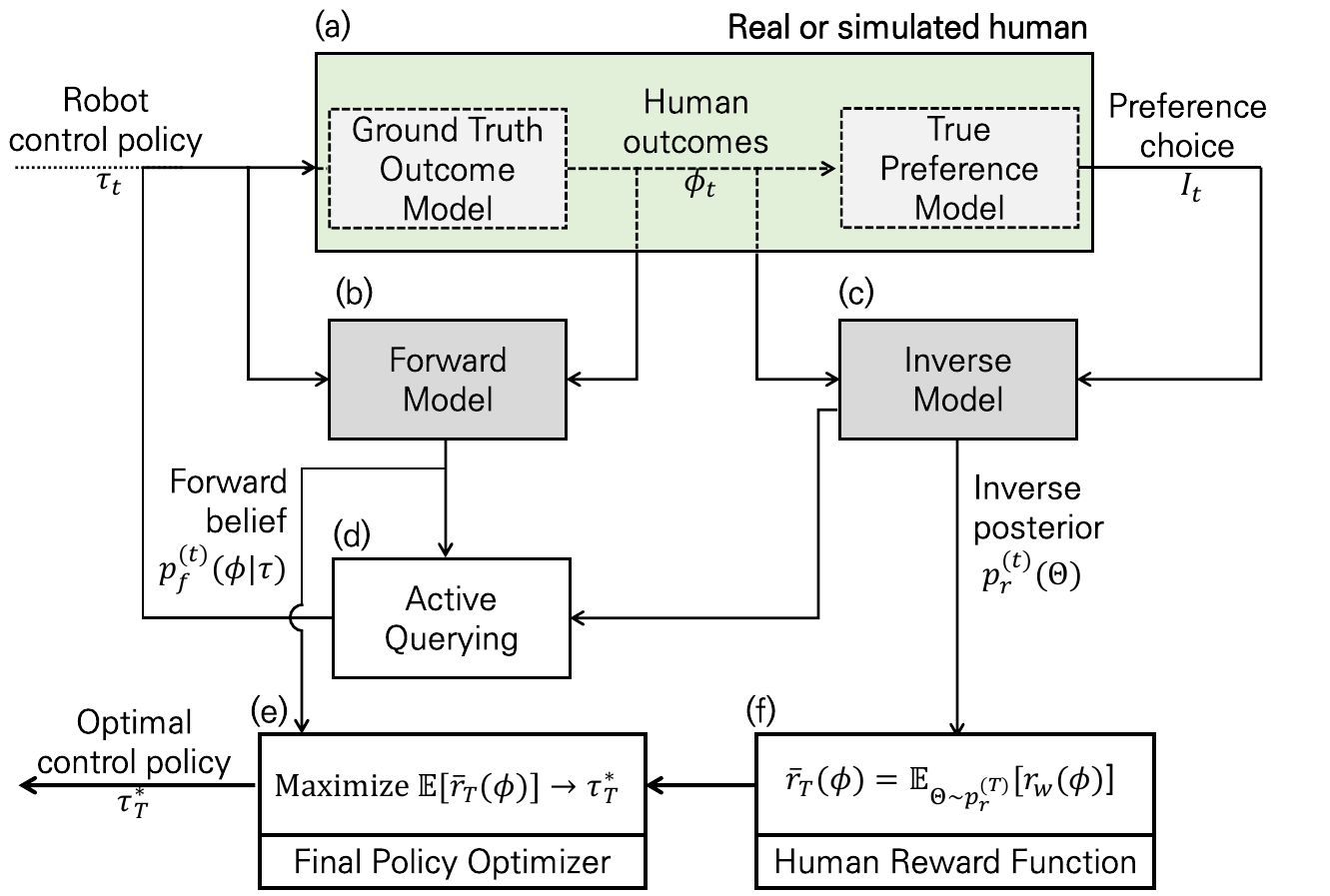}
\caption{
SFIHILO framework overview. (a) A real or simulated human compares a newly executed robot policy with the previous one; in simulation, this process is represented by a ground-truth outcome model and a true preference model. (b) The forward model learns a mapping from control policies to human outcomes. (c) The inverse model infers an outcome-space reward posterior from preference choices. (d) Active querying leverages both models to select informative policy comparisons. (e--f) After querying, the learned reward is used for final policy optimization.
}
    \label{fig:sfihilo_overview}
\end{figure*}

In this study, we propose \textit{Simultaneous Forward and Inverse Human-In-the-Loop Optimization} (\mbox{SFIHILO}) (Fig.~\ref{fig:sfihilo_overview}), which efficiently infers individual-specific reward functions from preferences over human outcomes and concurrently identifies a final control policy that maximizes the learned reward. We bootstrap a forward model that maps control policies to human outcomes; this model provides a predictive link for evaluating query effects, synthesizing informative queries for inverse reward learning, and selecting a final control policy to align robot behavior with the learned reward. Because SFIHILO involves a joint forward--inverse learning process, we propose a querying method that scores candidate control policies by their expected reduction in uncertainty across forward and inverse beliefs over a target preference-boundary region, reducing the required number of user trials.

To establish the algorithm's robustness before applying it to real users, we demonstrate in simulation that SFIHILO is effective across user heterogeneity, outcome dimensionalities, precision requirements, noise levels, and nonstationarity, with improved sample efficiency compared to mutual-information active-querying methods (Fig.~\ref{fig:final_results}). We also stress-test the algorithm under feature misspecification, providing guidance for feature selection in future human experiments. In summary, SFIHILO provides a new human-in-the-loop learning framework to (i) infer latent human rewards defined over human outcomes, (ii) learn a forward mapping from robot policies to human outcomes, (iii) actively synthesize user queries to optimize the simultaneous learning process, and (iv) align robot behavior with the learned human rewards. Code and data are available on \href{https://github.com/kyeongwon-park/SFIHILO}{GitHub}.

\section{Related Work}
Inverse reinforcement learning infers latent objectives from expert demonstrations~\cite{ng2000algorithms, abbeel2004apprenticeship, ziebart2008maximum, palan2019learning}, while reward learning extends to other forms of human supervision, including pairwise comparisons~\cite{christiano2017deep, pmlr-v87-biyik18a}, ratings~\cite{somers2017efficient, pmlr-v164-wilde22a}, rankings~\cite{myers2022learning}, and physical corrections~\cite{bajcsy2017learning}. These formulations are well suited to Markov decision processes where rewards are defined over states, actions, or trajectories~\cite{kim2021reward, brown2019extrapolating}, or more generally to settings in which options are representable by known reward-relevant features~\cite{sadigh2017active}. However, tracking fine-grained human states is often impractical due to partial observability, individual variability, and nonstationarity. Prior work on human--robot systems has therefore relied on cumulative or episodic outcomes, such as steady-state metabolic cost or fatigue of transport, to summarize policy effects~\cite{zhang2017human, ding2018human, kim2017human}, consistent with objective-level accounts of voluntary human movement~\cite{falisse2019rapid, carlisle2023optimization}. We likewise represent reward features as cumulative outcomes rather than state-transition trajectories. Crucially, as these outcomes are only revealed post-execution, forward modeling of the policy-to-outcome mapping is necessary for effective reward learning.

Forward models reduce experimental costs by predicting user outcomes before physical trials, serving as surrogates for costly policy-to-outcome evaluations, as in Bayesian optimization~\cite{shahriari2016loop}. For instance, real-time metabolic expenditure estimation from human limb motion~\cite{slade2021sensing} has been integrated into closed-loop exoskeleton personalization, avoiding the need for specialized equipment~\cite{slade2022personalizing}. More generally, in regression settings where observations are expensive and predictive uncertainty matters, uncertainty-aware regression is instrumental~\cite{rasmussen2006gaussian}. We build on these approaches to learn a predictive belief over a multidimensional mapping from control policies to outcomes.

For active inverse reward learning, prior work has either utilized amortized query generation from offline data collection~\cite{lee2023user} or maintained Bayesian reward posteriors~\cite{biyik2022aprel, biyik2024active, katz2021preference}. Most Bayesian reward models are linear in features~\cite{pmlr-v87-biyik18a, biyik2022aprel, katz2021preference}, though nonlinear extensions exist via post hoc neural feature augmentation~\cite{katz2021preference} or Gaussian processes~\cite{biyik2024active}; here, we employ Gaussian processes for inverse learning to efficiently handle potential nonlinearities in human rewards. Existing acquisition functions for active querying, including volume removal~\cite{sadigh2017active}, disagreement~\cite{katz2021preference}, and mutual information~\cite{pmlr-v87-biyik18a}, primarily focus on reward-side informativeness. Recently, expected value of information has been used to select queries by their anticipated downstream action improvements~\cite{myers2023active}. In this framework, we define the downstream objective as preference-boundary refinement. Because this boundary is defined over latent outcomes, resolving it requires joint lookahead across both the forward policy-to-outcome belief and the inverse outcome-to-reward belief. We propose a new criterion that scores policies by their expected reduction in uncertainty over the unresolved preference boundary, favoring queries with broad utility across the coupled forward--inverse learning problem.

Unlike standard reinforcement learning from human feedback (RLHF) pipelines that first learn a reward model from collected preferences and then optimize a policy against it~\cite{christiano2017deep,ouyang2022training,stiennon2020learning}, SFIHILO learns the forward and inverse models simultaneously and interactively under a limited online query budget, rather than from large offline datasets. Each query requires physically executing a policy on an individual user, so amortization across users or datasets is not available. SFIHILO cannot rely on a fixed pretrained forward model, because the mapping is user-specific and potentially nonstationary.

\section{Proposed Methods: SFIHILO}
\label{sec:methodology}

\paragraph{Human outcome and preference generation (Fig.~\ref{fig:sfihilo_overview}a).}
SFIHILO proceeds in a sequential loop (Fig.~\ref{fig:sfihilo_overview}). At round \(t\), SFIHILO executes a candidate control policy parameterized by \(\tau_t\in\mathcal{T}\subset\mathbb{R}^{d_\tau}\), which then serves as the anchor for the next round. The interaction structure is
\begin{equation}
    \tau \xrightarrow{\;f_{\star,t}\;} \phi \xrightarrow{\;r_\star\;} r_\star(\phi),
    \qquad
    f_{\star,t}:\mathcal{T}\rightarrow\Phi,
    \qquad
    r_\star:\Phi\rightarrow\mathbb{R},
    \label{eq:true_forward_map}
\end{equation}
where \(\phi_t=f_{\star,t}(\tau_t)\in\Phi\subset\mathbb{R}^{d_\phi}\) denotes the realized human outcome. The map \(f_{\star,t}\) captures the policy-to-outcome mapping, and \(r_\star\) denotes the user's latent reward function over outcomes. SFIHILO does not know either map a priori; it observes outcomes after execution and uses preference feedback over those outcomes to learn the reward structure. While \(f_{\star,t}\) can be treated as stationary in many settings, some users may adapt their responses to robot actions through motor learning (e.g., novice users)~\cite{poggensee2021adaptation}. We therefore allow \(f_{\star,t}\) to vary across rounds to model nonstationarity.

Each query compares a newly executed candidate policy $\tau_t$ with an anchor policy $\bar\tau_t$ whose realized outcome $\bar\phi_t$ has already been measured, so one new policy execution yields one comparison, following~\cite{lee2023user}. After $t$ completed preference queries, the histories are recorded as $\mathcal{D}_f^{(t)}=\{(\tau_i,\phi_i)\}_{i=1}^{t}$ and $\mathcal{D}_r^{(t)}=\{(\bar\phi_i,\phi_i,I_i)\}_{i=1}^{t}$, where $I_i\in\{+1,-1\}$ is the binary preference label for the ordered pair $(\bar\phi_i,\phi_i)$. SFIHILO uses these histories to maintain (i) a forward predictive belief $p_f^{(t)}(\phi\mid\tau)$ and (ii) an inverse posterior $p_r^{(t)}(\Theta)$, which are used in active querying and final policy selection.

\paragraph{Forward modeling (Fig.~\ref{fig:sfihilo_overview}b).}
The forward belief serves two roles: predicting the outcome of an unexecuted policy and identifying where uncertainty is experimentally reducible for active querying. After each new policy--outcome observation, the forward model is refit using the current history $\mathcal{D}_f^{(t)}$, maintaining a Gaussian predictive approximation,
\begin{equation}
    p_f^{(t)}(\phi\mid\tau)=\mathcal{N}\!\left(\mu_f^{(t)}(\tau),\,\Sigma_f^{(t)}(\tau)\right),
    \qquad
    \Sigma_f^{(t)}(\tau)=\mathrm{diag}\!\big(v_{f,1}^{(t)}(\tau),\dots,v_{f,d_\phi}^{(t)}(\tau)\big).
\end{equation}
Under kernel ridge regression (KRR), \(\mu_f^{(t)}(\tau)\) is the KRR predictive mean~\cite{saunders1998ridge} and each \(v_{f,j}^{(t)}(\tau)\) combines a leverage-based epistemic variance proxy with an observation-noise floor (Appendix~\ref{app:forward_krr}). For nonstationary settings, we use recency-weighted KRR with periodic hyperparameter reselection to adapt the effective recency weighting to individual motor-adaptation rates~\cite{klinkenberg2004learning} (Appendix~\ref{app:recency_weighted_forward}).

\paragraph{Inverse reward learning (Fig.~\ref{fig:sfihilo_overview}c).}
The inverse model maintains a posterior belief over the user's latent reward function on the outcome space. Its role is to explain the preference data \(\mathcal{D}_r^{(t)}\) in terms of reward-relevant outcomes and to provide uncertainty for active querying. We follow Gaussian-process preference learning~\cite{biyik2024active} to infer a nonlinear outcome-space reward from pairwise comparisons. To obtain a finite-dimensional representation for posterior inference, we approximate the Gaussian-process reward prior with a radial basis function (RBF) kernel using random Fourier features~\cite{rahimi2007random}.

Specifically, we place a Gaussian-process prior with an RBF kernel on the reward and a Gaussian prior on the log-choice sharpness \(\eta\), with \(\beta=e^\eta>0\):
\begin{equation}
    r \sim \mathcal{GP}(0, k_r), \quad
    \eta \sim \mathcal{N}(\mu_\eta, \sigma_\eta^2), \quad
    r(\phi) \approx r_w(\phi) = w^\top z(\phi), \;\;
    w \sim \mathcal{N}(0, \sigma_w^2 I_m).
    \label{eq:rff_reward_model}
\end{equation}
Here, \(k_r:\Phi\times\Phi\rightarrow\mathbb{R}\) is the outcome-space reward kernel, \(z:\Phi\rightarrow\mathbb{R}^m\) is the random Fourier feature map, and \(w\in\mathbb{R}^m\) denotes the finite-dimensional reward coefficients.
Letting \(\Theta=(w,\eta)\), each comparison induces the pairwise likelihood~\cite{christiano2017deep, sadigh2017active} as
\begin{equation}
    p(y \mid \phi_a, \phi_b, \Theta)
    =
    \sigma \!\left(
    y \beta \big( r_w(\phi_a) - r_w(\phi_b) \big)
    \right),
    \quad
p_r^{(t)}(\Theta) \propto p_0(\Theta)
\prod_{i=1}^{t}
p(I_i \mid \bar{\phi}_i,\phi_i,\Theta).
    \label{eq:preference_likelihood}
\end{equation}
where \(y\in\{+1,-1\}\) is a generic label value, \(\sigma(\cdot)\) is the standard logistic function, and \(p_0(\Theta)\) denotes the joint prior. The product uses the observed labels \(I_i\) from \(\mathcal{D}_r^{(t)}\). Because the logistic likelihood is non-conjugate, we apply a Laplace approximation around the posterior mode and draw posterior samples from the resulting Gaussian approximation~\cite{chu2005preference}.

These posterior samples define SFIHILO's posterior predictive preference distribution for any ordered outcome pair:
\begin{equation}
    q_t(y\mid \phi_a,\phi_b)
    =
    \mathbb{E}_{\Theta\sim p_r^{(t)}}
    \!\left[
    p(y\mid \phi_a,\phi_b,\Theta)
    \right],
    \qquad y\in\{+1,-1\}.
    \label{eq:reward_posterior_predictive}
\end{equation}
Here \(p(y\mid \phi_a,\phi_b,\Theta)\) denotes the likelihood term in Eq.~\eqref{eq:preference_likelihood} applied to \((\phi_a,\phi_b)\). We use \(q_t\) both to predict user preferences and as the inverse-model primitive in active querying.

\paragraph{Active querying (Fig.~\ref{fig:sfihilo_overview}d).}
SFIHILO chooses an executable policy, not an arbitrary outcome comparison. This distinction is central: a reward-informative comparison may be unreachable under the policy-to-outcome map, or may be reachable only imprecisely while the forward belief is uncertain. We therefore start from mutual information as the inverse-learning primitive, but value a policy query by how its observation changes future reward-informative querying. Let \(\mathcal{D}_t=(\mathcal{D}_f^{(t)},\mathcal{D}_r^{(t)})\), and let \(Y(\phi_a,\phi_b)\in\{+1,-1\}\) denote the hypothetical preference label for the outcome pair \((\phi_a,\phi_b)\).

\begin{enumerate}[leftmargin=*, nosep]
    \item \textbf{Mutual information baseline.}
    For a known outcome pair, define the outcome-space mutual-information score
    \begin{equation}
        \mathcal I\!\left(Y(\phi_a,\phi_b);\Theta\mid \phi_a,\phi_b,\mathcal{D}_t\right)
        =
        H\!\left(q_t(\cdot\mid\phi_a,\phi_b)\right)
        -
        \mathbb{E}_{\Theta\sim p_r^{(t)}}
        \!\left[
        H\!\left(p(\cdot\mid\phi_a,\phi_b,\Theta)\right)
        \right],
        \label{eq:mi_baseline}
    \end{equation}
    where \(H\) is binary entropy~\cite{houlsby2011bayesian}. The corresponding policy-space mutual-information baseline instantiates this score at the current forward mean, $\phi_a = \bar\phi_t$ and $\phi_b = \mu_f^{(t)}(\tau).$ This baseline is reward-informative, but uncertainty in the
forward belief can make the predicted informative outcome pair
difficult to realize reliably, highlighting the value of
forward learning.

    \item \textbf{Forward-valued lookahead.} Forward learning is therefore essential, but not as an objective for indiscriminate exploration (e.g., a standalone forward-variance bonus). Forward information is valuable insofar as it improves downstream reward learning. Instead of directly attempting to realize the outcome pair favored by the reward posterior, SFIHILO evaluates the one-step consequence of a candidate policy: predict possible observations (i.e., the outcome and preference label), update both the forward and inverse beliefs under each observation, and measure the resulting reduction in mutual information to assess the resolution of sample disagreements.

    \item \textbf{Target set for generalization.}
The key question is where to measure the expected reduction in mutual information. We do not measure this reduction only at the queried policy itself; once executed, that policy becomes part of the training history, so the resulting decrease in uncertainty would mainly reflect local assimilation rather than generalization. To assess generalization, the lookahead is evaluated on a finite target set \(\mathcal{Z}_t\subset\mathcal{T}\) of feasible policies, which approximates the current anchor-relative preference-boundary region. Specifically, $\mathcal{Z}_t$ comprises the top-ranked policies from a Monte Carlo pool with reward-informative predicted outcomes and experimentally reducible forward uncertainty (see Appendix~\ref{app:active_querying} for details).

    \item \textbf{Boundary Lookahead acquisition.}
For a target policy \(\zeta\in\mathcal{Z}_t\), let \(\Phi_\zeta\sim p_f^{(t)}(\cdot\mid\zeta)\) denote its forward-predictive outcome. We define the target-region utility as the average mutual information over \(\mathcal{Z}_t\):
    \begin{equation}
        \mathcal{U}_t(\mathcal{Z}_t\mid\mathcal{D}_t)
        =
        \frac{1}{|\mathcal{Z}_t|}
        \sum_{\zeta\in\mathcal{Z}_t}
        \mathcal{I}\!\left(
        Y(\bar\phi_t,\Phi_\zeta);
        \Theta,\Phi_\zeta
        \mid
        \bar\phi_t,\mathcal{D}_t
        \right).
        \label{eq:target_region_utility}
    \end{equation}
For a candidate query policy \(\tau\), a hypothetical observation is \(\tilde o=(\phi,y)\), where \(\phi\) is the realized outcome and \(y\in\{+1,-1\}\) is the preference label relative to the anchor \(\bar\phi_t\). Its predictive law is
    \begin{equation}
        p_t(\tilde o\mid\tau,\mathcal{D}_t)
        =
        p_f^{(t)}(\phi\mid\tau)\,
        q_t(y\mid\bar\phi_t,\phi),
        \label{eq:query_observation_law}
    \end{equation}
Given a realization \(\tilde o=(\phi,y)\), the corresponding hypothetical updated history is
    \begin{equation}
        \mathcal{D}_{t+1}^{\tau,\tilde o}
        =
        \left(
        \mathcal{D}_f^{(t)}\cup\{(\tau,\phi)\},
        \mathcal{D}_r^{(t)}\cup\{(\bar\phi_t,\phi,y)\}
        \right).
        \label{eq:hypothetical_update}
    \end{equation}
The Boundary Lookahead acquisition score is then defined as the expected reduction in the target-region utility after executing the candidate query and updating both beliefs:
    \begin{equation}
        a_{\mathrm{BL}}^{(t)}(\tau)
        =
        \mathcal{U}_t(\mathcal{Z}_t\mid\mathcal{D}_t)
        -
        \mathbb{E}_{\tilde o\sim p_t(\cdot\mid\tau,\mathcal{D}_t)}
        \!\left[
        \mathcal{U}_{t+1}\!
        \left(\mathcal{Z}_t\mid\mathcal{D}_{t+1}^{\tau,\tilde o}\right)
        \right].
        \label{eq:proposed_target_lookahead}
    \end{equation}
By design, this acquisition criterion can favor queries that are directly reward-informative, forward-informative for future targeting of reward-ambiguous regions, or both. For tractability, we approximate this lookahead using a finite candidate shortlist, a candidate mean-observation approximation, and finite forward-predictive sampling (see Appendix~\ref{app:active_querying}).

\end{enumerate}

\paragraph{Final policy selection (Fig.~\ref{fig:sfihilo_overview}e,f).} 
After \(T\) trials, we summarize the learned human reward by its posterior mean, 
\(\bar{r}_T(\phi) = \mathbb{E}_{\Theta \sim p_r^{(T)}} [r_w(\phi)]\). 
The final policy is then selected by maximizing the posterior-expected reward under the learned forward belief:
\begin{equation}
    \tau_T^\star = \arg\max_{\tau \in \mathcal{T}} 
    \mathbb{E}_{\phi \sim p_f^{(T)}(\cdot \mid \tau)} 
    \left[ \bar{r}_T(\phi) \right].
    \label{eq:final_policy_optimization}
\end{equation}

\section{Simulation Experiments}
\label{sec:experiments}

\subsection{Environment setup}\label{sec:settings} 
To establish algorithmic robustness prior to deployment, we systematically evaluate SFIHILO in simulation under conditions that capture the core challenges of assistive personalization: unknown policy-to-outcome mappings, nonlinear individual rewards, motor adaptation, and stochasticity.

\textbf{User heterogeneity.} Relevant human outcomes are aggregate, steady-state quantities that often vary smoothly with task or assistance parameters and exhibit few local minima~\cite{galle2017reducing, malcolm2013simple}. Accordingly, in the default setting, we sample the outcome and reward mappings of heterogeneous users, \(f_\star\) and \(r_\star\), from Gaussian processes with RBF kernels. By sampling entire functions rather than linear weights, this nonparametric approach directly captures nonlinear interactions and diverse, nonseparable trade-offs beyond simple additive preferences. To enable statistical validation, we sample 30 independent users and assess significance using Holm-corrected two-sided paired t-tests~\cite{holm1979simple}.

\textbf{Motor adaptation.} Following \cite{poggensee2021adaptation}, we model motor adaptation as a gradual exponential process that interpolates the policy-to-outcome mapping from an initial map to a steady-state map over query exposure. Empirical human--exoskeleton coadaptation has been reported to stabilize in approximately $330~\text{min}$ \cite{poggensee2021adaptation}. We set the time constant to achieve a 95\% transition by query 200 (time constant $\approx 67$). Assuming $2~\text{min}$ of interaction per query, our $400~\text{min}$ cumulative exposure is a more conservative choice than the experimental timescale reported in the literature.

\textbf{Stochastic execution and feedback.} Unless explicitly ablated, our simulations all incorporate realistic execution jitter, outcome measurement noise, and reward-evaluation noise (see Appendix~\ref{app:noise}).

\subsection{Evaluation protocol} 
We assessed the performance of SFIHILO by independently evaluating its two primary components. To quantify the continuous learning efficiency of the forward mapping, we report the area under the root-mean-square error curve (RMSE AUC). RMSE was computed with respect to the clean oracle outcome mean, $f_\star$, across randomly sampled control policies. The inverse model was assessed using posterior-predictive preference error on held-out outcome pairs. Because preference accuracy is intrinsically margin-dependent~\cite{lambert2025rewardbench,malik2026rewardbench2, wilde2023rightmeasure}, test pairs were stratified by the true reward gap $\Delta r = |r_\star(\phi_a)-r_\star(\phi_b)|$ into four bins: $[0.02,0.04)$, $[0.04,0.08)$, $[0.08,0.16)$, and $[0.16,\infty)$. A balanced aggregate set was used to ensure that performance was not dominated by large-margin comparisons. To avoid evaluating nearly indistinguishable outcomes, each pair was required to differ by at least 5\% of the full scale in at least one outcome dimension (i.e., a just-noticeable difference~\cite{gescheider1997psychophysics,prins2016psychophysics}).

While AUC characterizes refinement of the forward model, inverse-model sample efficiency is reported as the query count required to reach a 10\% preference-error threshold. For runs that did not reach this threshold, the count was extrapolated using a logarithmic fit. This 10\% threshold serves as a simulation benchmark, motivated by pairwise-consistency conventions in decision procedures~\cite{saaty1987analytic}; in physical studies, however, the practical error floor should be calibrated to human preference reliability, including inter-annotator disagreement and individual perceptual variability~\cite{ouyang2022training}. Because final policy selection in Eq. (\ref{eq:final_policy_optimization}) depends jointly on the learned forward belief and learned reward, we additionally report a final policy reward-gap diagnostic
(Fig.~\ref{fig:task_performance}; Appendix~A.9).

\begin{figure*}[t]
    \centering
    \includegraphics[width=1\textwidth]{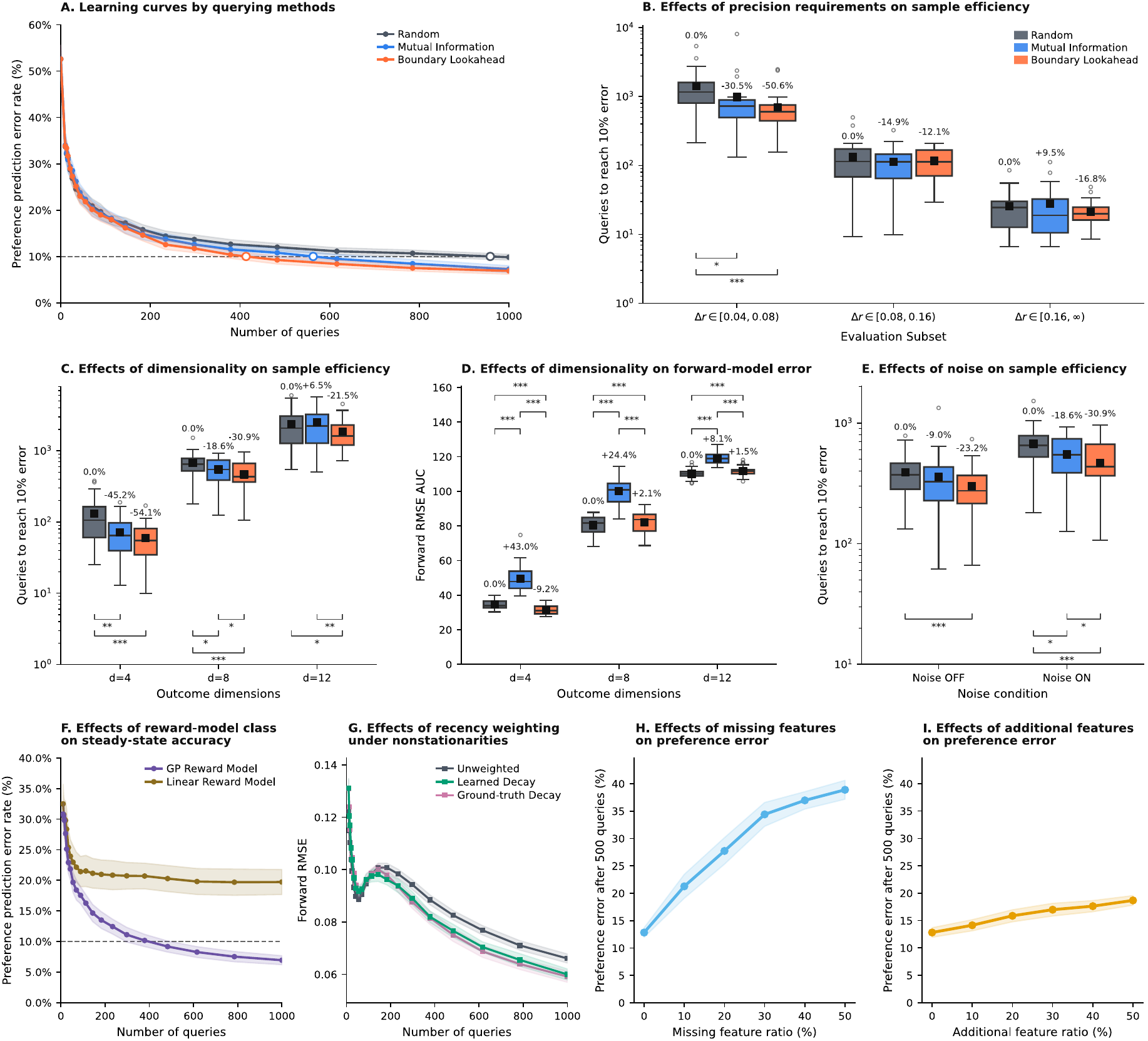}
\caption{
Simulation evaluation of SFIHILO across precision requirements, outcome dimensionalities, noise levels, nonstationarity, and feature misspecification. Unless varied along an axis, panels use the default settings: outcome dimensionality \(d=8\), practical noise levels, and reward-gap-balanced test sets; the feature-misspecification panels (H and I) use \(d=10\). Boxplots summarize independently sampled users. Annotations report the relative change from Random querying within each condition; asterisks denote Holm-adjusted paired t-tests~\cite{holm1979simple} (\(\ast\): \(p<0.05\), \(\ast\ast\): \(p<0.01\), \(\ast\ast\ast\): \(p<0.001\)).
}
\label{fig:final_results}
\end{figure*}

\subsection{Results}

\noindent\textbf{Active querying accelerated human reward modeling in simulation (Fig.~\ref{fig:final_results}A).} 
We evaluated the sample efficiency of SFIHILO by comparing Boundary Lookahead querying against Random querying and Mutual Information querying. Incorporating the forward model made it possible to implement active inverse learning in the simulated human--robot system. By linearly interpolating the mean learning curves on the reward-gap-balanced test set, we estimate that Random, Mutual Information, and Boundary Lookahead querying required 958, 563, and 413 queries, respectively, to reach a 10\% preference-error threshold. The 545-query reduction achieved by Boundary Lookahead relative to Random querying corresponds to 1,090 min, or 18.2 hours, of avoided human experimentation. The query budget depended strongly on the evaluation margin: on a moderate-difficulty test set ($|\Delta r|\geq 0.08$), Boundary Lookahead reached the same threshold after approximately 50 queries.

\noindent\textbf{Boundary Lookahead querying improved small-margin preference discrimination (Fig.~\ref{fig:final_results}B).}  
We tested the learned reward models on evaluation subsets stratified by the true reward gap, $\Delta r$, which indexes evaluation difficulty. Boundary Lookahead querying showed its largest relative benefit in the small-gap regimes, where preferences are hard to discriminate. This pattern is consistent with the method generating more informative data near the true preference boundary as the forward model became more accurate in reward-relevant regions. Notably, Mutual Information querying underperformed Random querying in the easiest, large-gap subset, whereas Boundary Lookahead querying retained its advantage. Together, these results suggest that Boundary Lookahead querying improves difficult preference-discrimination problems while preserving broad coverage of the reward landscape, rather than over-specializing to a narrow difficulty regime. The smallest reward-gap bin was excluded from query-count summaries because the 10\% crossing was not reliably estimable, although its learning curves qualitatively followed the same trend.

\noindent\textbf{Boundary Lookahead querying maintained sample efficiency as outcome dimensionality increased (Fig.~\ref{fig:final_results}C).} 
Prior studies on robot preference learning have frequently used low-dimensional reward features, whereas real-world human reward modeling may require higher-dimensional outcome spaces~\cite{falisse2019rapid}. This setting highlights the advantage of Boundary Lookahead querying under increasing outcome dimensionality. Mutual Information querying reduced the query count in low-dimensional settings but failed to maintain this benefit as dimensionality increased, eventually requiring more queries than Random querying at the highest dimensionality tested. In contrast, Boundary Lookahead querying maintained its sample efficiency across the dimensionality range considered here. We interpret this robustness as reflecting its boundary-region scoring rule, which targets general preference-learning improvement near unresolved preference boundaries, whereas Mutual Information querying can be drawn to posterior disagreement arising from competing extrapolations in weakly constrained regions of high-dimensional outcome spaces. These results suggest that Boundary Lookahead querying may provide practical benefits at the outcome dimensionalities expected in real applications.

\noindent\textbf{Boundary Lookahead querying improved inverse learning without sacrificing the forward model (Fig.~\ref{fig:final_results}D).}
SFIHILO uses the forward model for both query synthesis and final policy optimization, so gains in inverse-learning sample efficiency should not compromise policy-to-outcome regression. We therefore compared forward RMSE AUC across querying methods. Mutual Information querying was less effective at preserving forward accuracy, whereas Boundary Lookahead querying matched Random querying, which is expected to provide the broadest forward coverage. This accuracy helps translate reward-posterior preferences into feasible policies, supporting the learning efficiency observed above.

\noindent\textbf{Noise amplified the sample-efficiency gains from active querying (Fig.~\ref{fig:final_results}E).}
Noise increased absolute query counts but amplified the relative effectiveness of active querying. Mutual Information querying benefited by separating reducible preference uncertainty from aleatoric uncertainty, and Boundary Lookahead querying inherited this robustness through the mutual-information component of its acquisition score.

\noindent\textbf{Reward-model expressivity was necessary for high-precision reward identification (Fig.~\ref{fig:final_results}F).}
Under Boundary Lookahead querying, the Gaussian process reward model showed continued improvement, whereas the linear model saturated above the 10\% preference-error threshold. This saturation highlights the representational limits in capturing ground-truth nonlinearities, confirming that the efficacy of active querying is fundamentally bounded by the expressivity of the underlying reward model. The Gaussian process model may be a more flexible and robust choice for capturing real human preference landscapes.

\noindent\textbf{SFIHILO addressed nonstationarity through learned recency weighting (Fig.~\ref{fig:final_results}G).}
Motor adaptation can gradually shift users' biomechanical responses to robot actions, so SFIHILO uses recency-weighted forward modeling to track outcome drift. Because the forward model learned from noisy outcome observations, RMSE was partly noise-limited, making absolute differences between variants modest. However, learned decay closely tracked the ground-truth-decay reference and reduced late-stage forward RMSE relative to the unweighted baseline. This indicates that SFIHILO can adapt to simulated user-specific motor adaptation rates using the temporally ordered query history.

\noindent\textbf{Missing reward-relevant outcomes were more harmful than extra outcomes (Fig.~\ref{fig:final_results}H,I).}
The framework was evaluated under imperfect knowledge of the outcome features available to the inverse model, a frequent challenge in robot reward learning where the ground-truth reward-relevant feature set is unknown or misspecified~\cite{bobu2018learning,tien2023causal}. Missing features sharply increased preference error, indicating that omitted reward-relevant outcomes directly limit reward identifiability. Extra outcomes caused modest degradation, suggesting that the fixed query budget became less efficient as the learner's outcome space expanded. These asymmetric penalties suggest a practical feature-selection heuristic for applying SFIHILO in physical experiments: features that are moderately plausible as reward-relevant should generally be retained, whereas weakly motivated measurements should not be indiscriminately added to the learner's outcome space. After such controlled feature-set selection, primary reward contributors might be isolated through post hoc attribution or follow-up experiments.

\begin{table}[h]
\centering
\small
\setlength{\tabcolsep}{4pt}
\caption{Robustness across simulated-user families after 1,000 queries
(mean $\pm$ 95\% CI; $n=30$).}
\label{tab:reward_robustness}

\resizebox{0.92\linewidth}{!}{%
\begin{tabularx}{\linewidth}{@{}l*{3}{>{\centering\arraybackslash}X}@{}}
\toprule
\shortstack[l]{Simulated-user\\family}
& \shortstack{Forward\\RMSE $\downarrow$}
& \shortstack{Preference\\error (\%) $\downarrow$}
& \shortstack{Final reward gap\\to optimum $\downarrow$} \\
\midrule
RBF (matched)
& $0.0635 \pm 0.0024$
& $6.88 \pm 0.72$
& $0.0630 \pm 0.0224$ \\
Matérn-$5/2$
& $0.0473 \pm 0.0011$
& $6.83 \pm 1.16$
& $0.0496 \pm 0.0189$ \\
Additive polynomial
& $0.0411 \pm 0.0004$
& $6.39 \pm 0.64$
& $0.0864 \pm 0.0321$ \\
Random MLP
& $0.0535 \pm 0.0010$
& $7.67 \pm 0.68$
& $0.0808 \pm 0.0227$ \\
\bottomrule
\end{tabularx}%
}
\end{table}

\noindent\textbf{SFIHILO remained effective with non-RBF simulated users (Table~\ref{tab:reward_robustness}).} We used Gaussian process priors with RBF kernels as a smooth, stationary reference for gradual human responses. To test dependence on this matched setting, we kept SFIHILO's RBF-based forward and inverse learners fixed and evaluated them across three non-RBF ground-truth families: (1) a Matérn-$5/2$ Gaussian process with finite smoothness and rougher local variation; (2) an additive low-order polynomial with global linear-to-cubic structure; and (3) a random two-hidden-layer multilayer perceptron (MLP) with flexible, ridge-like nonlinearities. Held-out complexity diagnostics showed no gross family-level imbalance in sampled-function difficulty (Appendix ~\ref{app:non_rbf_families}). Across all four families, SFIHILO maintained comparable forward RMSE, preference error, and final reward gap, supporting robustness to family mismatch, which may facilitate transfer to real users.

\begin{figure*}[h]
    \centering
    \includegraphics[width=0.72\textwidth]{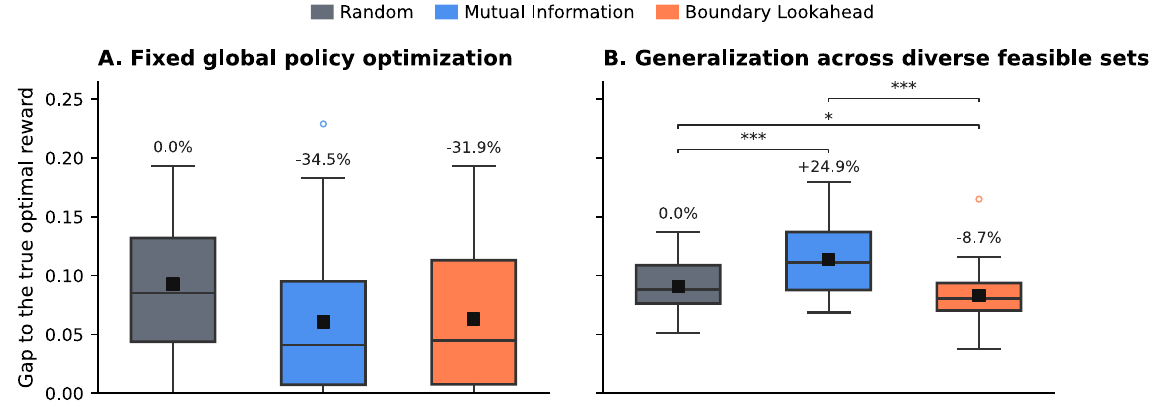}
    \caption{
    Final reward gap to the optimum after 1,000 preference queries under
    (A) optimization over the fixed global policy space and
    (B) reoptimization across 210 balanced feasible sets
    ($\binom{8}{4}$ constrained-parameter subsets $\times$ three complementary
    low/mid/high assignments).
    }
    \label{fig:task_performance}
\end{figure*}

\noindent\textbf{Boundary Lookahead querying improved policy selection under new control constraints (Fig.~\ref{fig:task_performance}).} We evaluated final policy performance of SFIHILO in two policy-selection scenarios: optimization over a fixed global policy space and reoptimization across varying feasible sets. We expect the latter to be common in practice, because hardware characteristics and application-specific safety constraints can alter the feasible space across devices or control settings. In fixed global-policy optimization (Fig.~\ref{fig:task_performance}A), Boundary Lookahead querying performed comparably to Mutual Information querying, as expected because Mutual Information querying directly targets reward-side informativeness. When the learned model is instead reused across 210 different feasible sets, each restricting four of eight control parameters to prescribed ranges (Fig.~\ref{fig:task_performance}B), Boundary Lookahead achieved the lowest final reward gap. The proposed method showed more reliable policy reoptimization across diverse control constraints, supporting the flexibility needed for transfer across deployments.

\section{Limitations and Future Work}
SFIHILO has yet to be validated in physical human trials. We focused on algorithm development and validation through controlled ablations in simulation, which provides access to ground-truth human reward functions and policy-to-outcome maps. For real humans, these quantities are often unobservable, and algorithmic error is difficult to disentangle from dimensionality, precision requirements, noise, user adaptation, and feature misspecification. Successful deployment will depend on an appropriate outcome representation, control-policy design that selectively varies those outcomes, and reliable sensing. Future work should develop benchmarks that distinguish levels of discrimination difficulty in real users and assess whether the final policy improves their experience.


\clearpage


\bibliography{main}  

\clearpage

\appendix
\section{Appendix}
\label{sec:appendix}

\paragraph{Supplementary code and data.}
An implementation, smoke-test scripts, precomputed simulation data, and plotting scripts are available at \url{https://github.com/kyeongwon-park/SFIHILO}.

\subsection{Simulated-user generation}
\label{app:simulated_user_generation}

\begin{figure*}[h!]
    \centering
    \begin{minipage}[t]{0.675\textwidth}
        \vspace{0pt}
        \centering
        \includegraphics[width=0.9\linewidth]{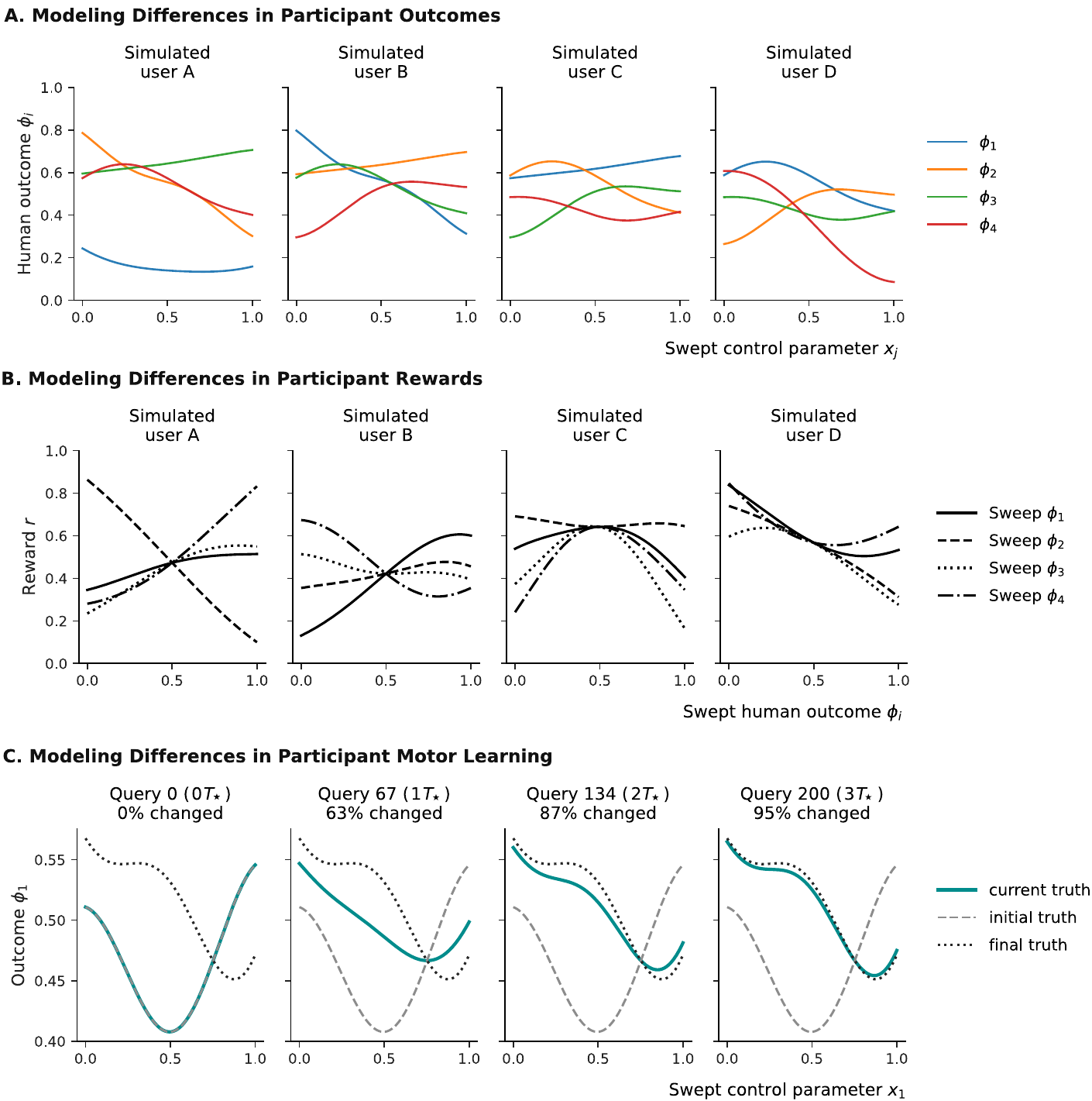}
    \end{minipage}
    \hfill
    \begin{minipage}[t]{0.315\textwidth}
        \vspace{4pt}
        \caption{
        Model-behavior diagnostics for simulated users.
        (A) Policy-to-outcome sweeps for four independently sampled simulated users, showing smooth and diverse outcome responses to control variation.
        (B) Outcome-to-reward sweeps for the same users, showing smooth nonlinear preference structure over human outcomes.
        (C) Nonstationary outcome drift for motor adaptation, where the current policy-to-outcome map exponentially transitions from an initial response map to a steady-state response map with time constant \(\approx67\) queries, reaching 95\% transition by query 200.
        }
        \label{fig:truth_behavior}
    \end{minipage}
\end{figure*}

Figure~\ref{fig:truth_behavior} shows model-behavior diagnostics for independently sampled simulated users. 
We use radial-basis-function Gaussian-process (RBF-GP) priors as a smooth, nonparametric family for generating simulated-user response functions. 
This choice follows the structure of the human-in-the-loop assistance problems considered in the main text: relevant outcomes are often cumulative or episodic summaries, such as steady-state or aggregate quantities, that vary continuously with task or assistance parameters and commonly exhibit few local minima~\cite{galle2017reducing,malcolm2013simple}. 
An RBF-GP prior provides an appropriate level of smoothness and curvature for these response surfaces while still allowing heterogeneous nonlinear effects and nonseparable interactions across policy dimensions. 
The same smoothness assumption is also a natural starting point for outcome-space rewards: absent discrete contextual switches or hidden thresholds, nearby human outcomes should generally induce nearby latent reward values. 
Accordingly, each simulated user was generated by sampling a latent policy-to-outcome map \(f_\star\) and a latent outcome-to-reward map \(r_\star\) using random Fourier feature (RFF) approximations to RBF-GP priors~\cite{rahimi2007random}. In the reported dimensionality sweeps, we set \(d_\tau=d_\phi=d\) so that
increasing the outcome dimensionality also increases the dimensionality of the
reachable policy-induced outcome manifold, rather than embedding many correlated
outcome coordinates in a fixed low-dimensional control space.

For a generic scalar RFF head with input \(x\in[0,1]^d\), we used
\begin{equation}
    g(x)
    =
    \sum_{m=1}^{M}
    w_m
    \sqrt{\frac{2}{M}}
    \cos\!\left(
    \frac{\omega_m^\top x}{\ell} + b_m
    \right),
    \quad
    \omega_m\sim\mathcal{N}(0,I),\;
    b_m\sim\mathrm{Unif}(0,2\pi),\;
    w_m\sim\mathcal{N}(0,1).
    \label{eq:truth_rff_head}
\end{equation}
The length scale for each sampled head was drawn log-uniformly,
\begin{equation}
    \ell
    =
    \exp\!\left((1-u)\log \ell_{\min} + u\log \ell_{\max}\right),
    \qquad
    u\sim\mathrm{Unif}(0,1).
    \label{eq:truth_rff_lengthscale}
\end{equation}

For the policy-to-outcome map, one independent RFF head was sampled for each outcome dimension, yielding raw outcomes \(\tilde f_{\star,j}(\tau)=g^{(f)}_j(\tau)\). 
These raw outputs were then min--max normalized to \([0,1]\) using calibration policies uniformly sampled from the control space, reflecting the normalization process used in practical multiobjective human-in-the-loop optimization experiments~\cite{lakmazaheri2024optimizing}. 
The outcome-to-reward map was sampled analogously as a scalar RFF function \(\tilde r_\star(\phi)=g^{(r)}(\phi)\) over the outcome space and then min--max normalized using outcomes induced by uniformly sampled control policies. 
The absolute scale of \(r_\star\) is not intrinsically identifiable from pairwise preferences~\cite{christiano2017deep, pmlr-v87-biyik18a}: the learner primarily observes the induced ranking of outcomes, with relative margins mattering through the choice-noise model. 
We therefore use this normalization to place rewards on comparable bounded scales for numerical stability and controlled evaluation margins, while preserving the nonlinear structure of each sampled function.

\begin{table}[h]
\centering
\small
\setlength{\tabcolsep}{4pt}
\caption{RFF hyperparameters used to sample simulated users. These settings define the simulated-user response functions, not the learned inverse reward model.}
\label{tab:simulated_user_rff}
\begin{tabularx}{\linewidth}{@{}l
>{\raggedright\arraybackslash}X
>{\raggedright\arraybackslash}X@{}}
\toprule
Quantity & Policy-to-outcome map \(f_\star\) & Outcome-to-reward map \(r_\star\) \\
\midrule
Input & Control policy \(\tau\in[0,1]^{d_\tau}\) & Outcome \(\phi\in[0,1]^{d_\phi}\) \\
Output & One scalar head per outcome dimension & One scalar reward head \\
RFF features & \(M_f=1024\) & \(M_r=512\) \\
Length scale & \(\ell_f\sim\mathrm{LogUniform}(0.35,0.65)\) & \(\ell_r\sim\mathrm{LogUniform}(0.50,0.80)\) \\
Frequencies & \(\omega_m\sim\mathcal{N}(0,I)\) & \(\omega_m\sim\mathcal{N}(0,I)\) \\
Phases & \(b_m\sim\mathrm{Unif}(0,2\pi)\) & \(b_m\sim\mathrm{Unif}(0,2\pi)\) \\
Weights & \(w_m\sim\mathcal{N}(0,1)\) & \(w_m\sim\mathcal{N}(0,1)\) \\
Head sharing & Independent heads across outcome dimensions & Single reward head \\
Calibration size & \(N_{\mathrm{cal},x}=20{,}000\) policies & \(N_{\mathrm{cal},\phi}=20{,}000\) induced outcomes \\
Normalization & Min--max scaling to \([0,1]\) & Min--max scaling to \([0,1]\) \\
\bottomrule
\end{tabularx}
\end{table}

Figure~\ref{fig:truth_behavior}C visualizes the nonstationary policy-to-outcome map used to model motor adaptation. 
Following~\cite{poggensee2021adaptation}, we model adaptation as a gradual change in the policy-to-outcome response. 
For each simulated user, we sample an initial map \(f_{\star,0}\) and a steady-state map \(f_{\star,\infty}\) using the RFF-GP procedure described above, and define the current map as
\begin{equation}
    f_{\star,t}(\tau)
    =
    \left(1-\alpha_\star(t)\right) f_{\star,0}(\tau)
    +
    \alpha_\star(t) f_{\star,\infty}(\tau),
    \qquad
    \alpha_\star(t)=1-\exp(-t/T_\star).
    \label{eq:app_motor_adaptation_oracle}
\end{equation}
Thus, nonstationarity affects how robot policies map to measured human outcomes, while the latent reward \(r_\star(\phi)\) remains defined on stable outcome coordinates. We set \(T_\star\) so that the transition reaches \(95\%\) by query \(t=200\),
\begin{equation}
    T_\star
    =
    -\frac{200}{\log(1-0.95)}
    \approx 66.8
    \quad\text{queries}.
    \label{eq:app_motor_adaptation_time_constant}
\end{equation}
Assuming \(2~\mathrm{min}\) of interaction per query, this corresponds to \(400~\mathrm{min}\) of cumulative exposure by the 95\% transition, a conservative timescale relative to the approximately \(330~\mathrm{min}\) stabilization reported for human--exoskeleton coadaptation~\cite{poggensee2021adaptation}.

\subsection{Non-RBF simulated-user families}
\label{app:non_rbf_families}

To evaluate robustness to ground-truth family mismatch, we generated additional simulated users from Mat\'ern-$5/2$, additive-polynomial, and nonlinear MLP families. For each family, we independently sampled eight policy-to-outcome heads and one outcome-to-reward head and applied the same calibration protocol used for the RBF--RFF reference. No optimum, concavity, or boundary structure was imposed.

The Mat\'ern-$5/2$ family used random Fourier features sampled from the corresponding spectral density, with the same feature counts as the RBF--RFF reference and both endpoints of the reference forward and reward length-scale intervals doubled. The additive-polynomial family summed normalized Legendre terms of degrees one through three across input dimensions, with relative coefficient scales \((0.50,1.00,0.75)\). The nonlinear MLP family used two tanh hidden layers of width 64, fan-in-normalized Gaussian weights, hidden-bias standard deviation \(0.50\), and gains of \(2.50\) and \(1.45\) for the forward and reward maps, respectively. The Mat\'ern length-scale factor and MLP gains were selected using pilot seeds disjoint from the 30 evaluation seeds.

To assess gross family-level differences in sampled-function difficulty, we evaluated the composed objective \(r_\star(f_\star(\tau))\) along 24 random policy chords, i.e., line segments connecting randomly sampled policy pairs, using 401 points per chord (Table~\ref{tab:truth_family_diagnostics}). For each seed, we computed the median chord-wise reward range and the fraction of chords containing at least one interior peak with prominence of \(0.02\) or greater. We additionally measured affine-residual nonlinearity as the fraction of variance unexplained by the best affine fit (\(1-R^2\)) over an independent set of 4{,}096 random policies. Each statistic was then averaged across the 30 evaluation seeds. Taken together, these diagnostics did not indicate a gross family-level imbalance in sampled-function difficulty, despite variation in affine-residual nonlinearity.

\begin{table}[h]
\centering
\small
\setlength{\tabcolsep}{4pt}
\caption{Ground-truth complexity diagnostics across simulated-user families. Values are averaged over 30 users evaluated with identical random-policy probes.}
\label{tab:truth_family_diagnostics}
\begin{tabularx}{\linewidth}{@{}l*{3}{>{\centering\arraybackslash}X}@{}}
\toprule
Truth family
& \shortstack{Median chord\\range}
& \shortstack{Fraction of chords\\with an interior peak}
& \shortstack{Affine-residual\\nonlinearity} \\
\midrule
RBF--RFF
& 0.203
& 0.486
& 0.826 \\
Mat\'ern-$5/2$
& 0.192
& 0.407
& 0.582 \\
Additive polynomial
& 0.179
& 0.424
& 0.902 \\
Random MLP
& 0.210
& 0.403
& 0.408 \\
\bottomrule
\end{tabularx}
\end{table}

\subsection{Simulation noise model} 
\label{app:noise}

All policy and outcome coordinates are normalized before adding noise, so the noise
levels can be interpreted as fractions of the full scale of each coordinate. At round \(t\),
the query rule selects a commanded policy \(\tau_t\in\mathcal T\). Execution variability
perturbs the implemented policy before the simulated policy-to-outcome map is evaluated:
\begin{equation}
    \tau_t^{\mathrm{exec}}
    =
    \Pi_{\mathcal T}\big(\tau_t+\epsilon_{\tau,t}\big),
    \qquad
    \epsilon_{\tau,t}\sim
    \mathcal N(0,\sigma_\tau^2 \mathbf I_{d_\tau}),
    \label{eq:app_execution_noise}
\end{equation}
where \(\Pi_{\mathcal T}\) denotes projection to the feasible policy domain. The clean
outcome \(\phi_t^\star=f_{\star,t}(\tau_t^{\mathrm{exec}})\) is then observed with sensing
noise:
\begin{equation}
    \phi_t
    =
    \Pi_{\Phi}\big(\phi_t^\star+\epsilon_{\phi,t}\big),
    \qquad
    \epsilon_{\phi,t}\sim
    \mathcal N(0,\sigma_\phi^2 \mathbf I_{d_\phi}),
    \label{eq:app_sensing_noise}
\end{equation}
where \(\Pi_{\Phi}\) denotes projection to the outcome domain. The commanded policy
and observed outcome are stored in the forward history.

Preference labels are generated by noisy evaluations of the latent reward. For an
ordered comparison \((\phi_a,\phi_b)\), the simulator forms
\begin{equation}
    \tilde r_a = r_\star(\phi_a)+\epsilon_{\tilde r,a},
    \qquad
    \tilde r_b = r_\star(\phi_b)+\epsilon_{\tilde r,b},
    \qquad
    \epsilon_{\tilde r,a},\epsilon_{\tilde r,b}
    \overset{\mathrm{i.i.d.}}{\sim}
    \mathcal N(0,\sigma_{\tilde r}^2),
    \label{eq:app_reward_eval_noise}
\end{equation}
and sets \(I_t=\operatorname{sign}(\tilde r_a-\tilde r_b)\). No additional label-flip
stage is used.

\begin{table}[h]
\centering
\caption{Default stochastic-noise settings. Standard deviations are in normalized units; FS denotes the full scale of a normalized coordinate.}
\label{tab:app_noise_settings}
\begin{tabularx}{\linewidth}{@{}l c X@{}}
\toprule
Noise source & Std. & Scale and role \\
\midrule
Execution 
& \(\sigma_\tau=0.02\) 
& 2\% FS implementation jitter in the commanded policy \\
Outcome measurement 
& \(\sigma_\phi=0.05\) 
& 5\% FS sensing noise in measured outcomes \\
Reward evaluation 
& \(\sigma_{\tilde r}=0.02\) 
& ambiguity for small reward-gap comparisons, with reversal rates determined by \(\Delta r\) \\
\bottomrule
\end{tabularx}
\end{table}

The reward-evaluation setting was chosen to make near-boundary comparisons noisy
without corrupting large-margin preferences. For a clean reward gap
\(\Delta r=|r_\star(\phi_a)-r_\star(\phi_b)|\), independent reward-evaluation noise
induces a noisy-gap standard deviation
\(\sigma_\Delta=\sqrt{2}\sigma_{\tilde r}\) and a noise-induced reversal probability
\begin{equation}
    P_{\mathrm{rev}}(\Delta r)
    =
    \Phi_{\mathcal N}
    \left(
    -\frac{\Delta r}{\sqrt{2}\sigma_{\tilde r}}
    \right),
    \label{eq:app_noise_reversal_prob}
\end{equation}
where \(\Phi_{\mathcal N}\) is the standard normal cumulative distribution function. With
\(\sigma_{\tilde r}=0.02\), this gives reversal probabilities of \(24.0\%\),
\(7.9\%\), \(0.23\%\), and \(7.7\times10^{-7}\%\) at
\(\Delta r=0.02,0.04,0.08,\) and \(0.16\), respectively. Thus, the model makes
difficult, small-gap comparisons substantially more likely to be confused, while
large-margin preferences remain effectively deterministic. This captures the expected
structure of human preference feedback, where near-indifferent alternatives are less
reliably distinguished than clearly separated ones.

These stochastic terms are applied to the active closed-loop interaction process to
reflect noise from both the physical environment and human feedback. Evaluation labels, however, are computed from the clean
underlying reward function \(r_\star\), rather than from stochastic reward-evaluation
samples. Thus, reported preference accuracy measures how well the learner recovers
the underlying preference landscape under noisy closed-loop interaction.

\subsection{Estimating inverse sample efficiency}
\label{app:evaluation}

\begin{figure*}[h!]
    \centering
    \includegraphics[width=\textwidth]{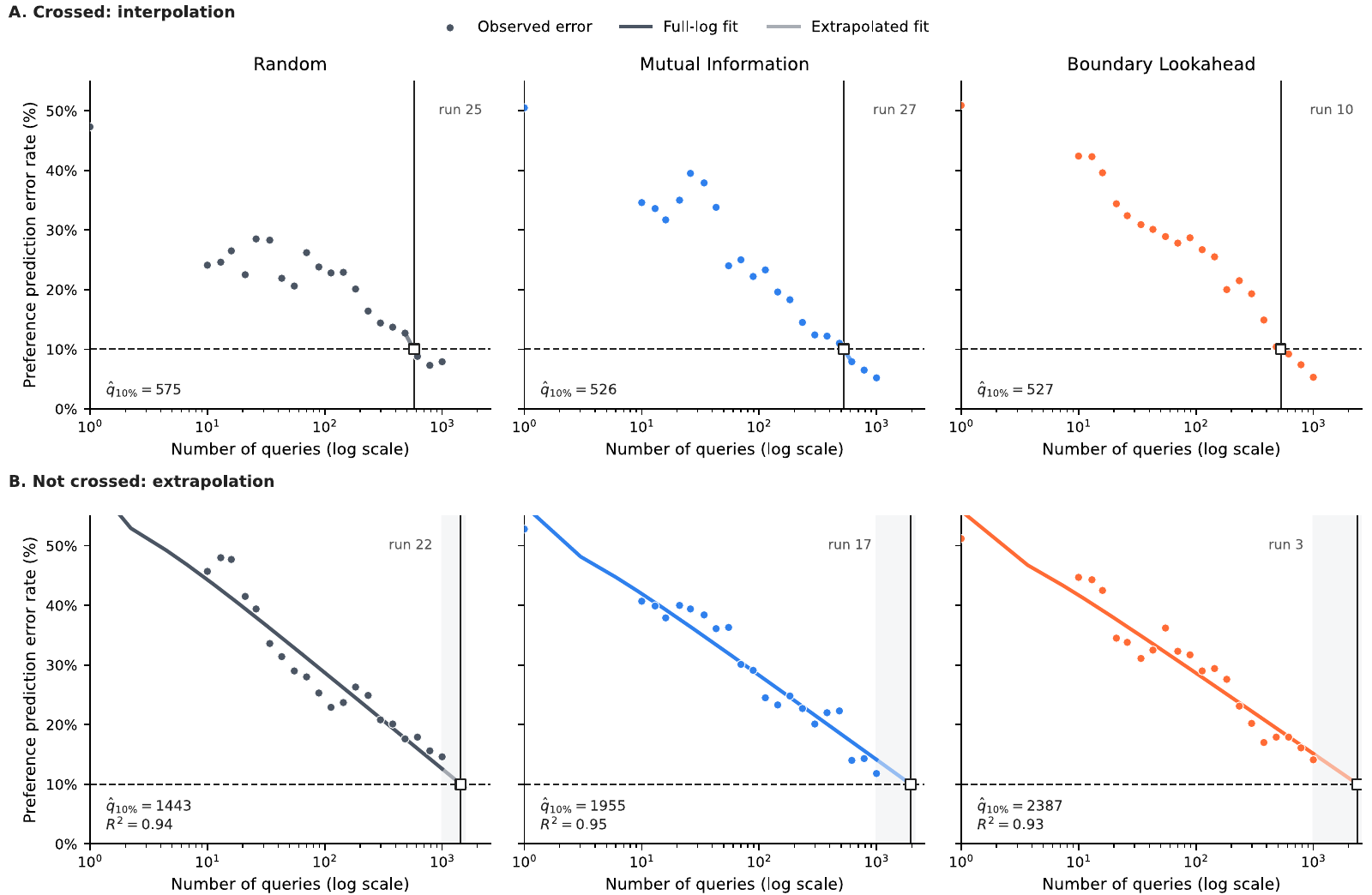}
\caption{
Estimation of the 10\% preference-error query threshold. Each column shows a
representative run for one querying rule. (A) When the observed held-out error
crosses 10\% within the finite query budget, the threshold is estimated by linear
interpolation between adjacent checkpoints; no curve fit is used. (B) Only for runs
that do not cross within \(q_{\max}=1000\) queries, we fit a logarithmic learning
curve to the observed checkpoints and solve for the crossing. Points denote observed
errors, dark curves denote the fitted log trend, and light curve segments denote the
extrapolated portion. The dashed horizontal line marks 10\% error, and square markers
indicate the estimated threshold \(\hat q_{10\%}\). The fitted examples show that the
logarithmic trend captures the observed decrease before extrapolation.
}
\label{fig:log_extrapolation}
\end{figure*}

We summarize inverse-learning sample efficiency by the number of active-learning
queries required for held-out posterior-predictive preference error to reach \(10\%\)
(Fig.~\ref{fig:log_extrapolation}). Let \(e(q)\) denote the fractional error after \(q\)
queries. Thresholds were estimated directly from observed checkpoints whenever the
held-out error crossed 10\% within the finite query budget \(q_{\max}=1000\).
Although this budget also keeps the simulation study computationally tractable, its
primary motivation is the practical user-burden constraint of physical human-in-the-loop
experiments: each query represents a commanded policy that a person must physically
experience, an outcome measurement that must be collected, and a preference comparison
that must be elicited. Thus, \(q_{\max}\) can be interpreted as a practical interaction
horizon. For these runs, if
\(e(q_i)>0.10\) and \(e(q_{i+1})\le 0.10\), we estimate the crossing by linear
interpolation in query count,
\begin{equation}
    \hat q_{10\%}
    =
    q_i
    +
    \frac{0.10-e(q_i)}
    {e(q_{i+1})-e(q_i)}
    \big(q_{i+1}-q_i\big).
    \label{eq:app_linear_crossing}
\end{equation}
Thus, the primary estimate is an observed crossing estimate rather than a fitted
extrapolation (Fig.~\ref{fig:log_extrapolation}A).

For the remaining non-crossing runs, whose observed learning curves do not reach the
target error within this practical interaction horizon, we fit the observed learning
curve with
\begin{equation}
    e(q)=a\log(1+q)+b,
    \qquad a<0,
    \label{eq:app_logfit_error}
\end{equation}
and solve for the 10\% crossing,
\begin{equation}
    \hat q_{10\%}
    =
    \exp\!\left(\frac{0.10-b}{a}\right)-1 .
    \label{eq:app_logfit_crossing}
\end{equation}
This log extrapolation is used only for threshold-based summaries of non-crossing
runs (Fig.~\ref{fig:log_extrapolation}B). All learning-curve panels report observed errors over the simulated query range.

\subsection{Forward KRR predictive belief}
\label{app:forward_krr}

The forward model is used to construct the predictive belief
\(p_f^{(t)}(\phi\mid\tau)\), mapping control policies to human outcomes. Given the forward history 
$\mathcal{D}_f^{(t)}=\{(\tau_i,\phi_i)\}_{i=1}^{t}$, we fit a multi-output RBF kernel ridge regressor (KRR)~\cite{saunders1998ridge}. Let
\begin{equation}
    k_{\ell_f}(\tau,\tau')
    =
    \exp\!\left(
    -\frac{\|\tau-\tau'\|_2^2}{2\ell_f^2}
    \right),
    \qquad
    K_t = [k_{\ell_f}(\tau_i,\tau_j)]_{i,j=1}^{t},
    \label{eq:app_forward_kernel}
\end{equation}
and define
\begin{equation}
    A_t = K_t + (\lambda_f+\epsilon_f)I .
    \label{eq:app_forward_kernel_system}
\end{equation}
Outcomes are standardized coordinatewise before fitting. Specifically, if \(m_t,s_t\in\mathbb R^{d_\phi}\) are the empirical outcome mean and standard deviation, then
\(\tilde\phi_{i,j}=(\phi_{i,j}-m_{t,j})/s_{t,j}\). Writing
\(\tilde{\Phi}_t=[\tilde\phi_1,\ldots,\tilde\phi_{t}]^\top\), the KRR coefficients are
\begin{equation}
    B_t = A_t^{-1}\tilde{\Phi}_t .
    \label{eq:app_forward_coefficients}
\end{equation}
For a new policy \(\tau\), let
\(k_t(\tau)=[k_{\ell_f}(\tau,\tau_1),\ldots,k_{\ell_f}(\tau,\tau_{t})]^\top\). The forward predictive mean is
\begin{equation}
    \mu_f^{(t)}(\tau)
    =
    m_t
    +
    s_t\odot
    \left(
    k_t(\tau)^\top B_t
    \right),
    \label{eq:app_forward_mean}
\end{equation}
where \(\odot\) denotes elementwise multiplication.

\paragraph{Predictive uncertainty decomposition.}
The forward model uses a Gaussian predictive approximation with diagonal covariance,
\begin{equation}
    p_f^{(t)}(\phi\mid\tau)
    =
    \mathcal N\!\left(
    \mu_f^{(t)}(\tau),
    \operatorname{diag}(v_f^{(t)}(\tau))
    \right).
    \label{eq:app_forward_predictive}
\end{equation}
The variance is decomposed into an experimentally reducible epistemic component and an irreducible observation component. The leverage-based epistemic factor is
\begin{equation}
    e_t(\tau)
    =
    \left[
    1
    -
    k_t(\tau)^\top A_t^{-1}k_t(\tau)
    \right]_+ ,
    \label{eq:app_forward_leverage}
\end{equation}
where \([\cdot]_+\) clips negative numerical artifacts to zero. Let
\(\hat\sigma_{\mathrm{resid},t,j}^2\) denote the residual variance for outcome coordinate \(j\), and let
\(\sigma_\phi^2\) denote the outcome measurement-noise variance. The coordinatewise epistemic scale is
\begin{equation}
    \gamma_{t,j}^2
    =
    \left[
    s_{t,j}^2-\hat\sigma_{\mathrm{resid},t,j}^2
    \right]_+ ,
    \label{eq:app_forward_epi_scale}
\end{equation}
and the variance decomposition is
\begin{equation}
    v_{\mathrm{epi},j}^{(t)}(\tau)
    =
    e_t(\tau)\gamma_{t,j}^2,
    \qquad
    v_{\mathrm{obs},j}^{(t)}
    =
    \hat\sigma_{\mathrm{resid},t,j}^2+\sigma_\phi^2,
    \qquad
    v_{f,j}^{(t)}(\tau)
    =
    v_{\mathrm{epi},j}^{(t)}(\tau)
    +
    v_{\mathrm{obs},j}^{(t)} .
    \label{eq:app_forward_variance_decomposition}
\end{equation}
This decomposition prevents active querying from valuing outcomes merely because they are noisy: only \(v_{\mathrm{epi},j}^{(t)}\) is treated as reducible through additional policy executions.

Forward samples used in downstream acquisition are drawn as
\begin{equation}
    \Phi_\tau^{(m)}
    \sim
    \operatorname{clip}_{[0,1]^{d_\phi}}
    \mathcal N\!\left(
    \mu_f^{(t)}(\tau),
    \operatorname{diag}(v_f^{(t)}(\tau))
    \right),
    \qquad
    m=1,\ldots,M_f .
    \label{eq:app_forward_sampling}
\end{equation}

\paragraph{Experimentally reducible uncertainty scores.}
The active-querying implementation uses two summaries of the reducible component. The first is the fraction of predictive uncertainty that is experimentally reducible,
\begin{equation}
    \rho_f^{(t)}(\tau)
    =
    \frac{1}{d_\phi}
    \sum_{j=1}^{d_\phi}
    \frac{
    v_{\mathrm{epi},j}^{(t)}(\tau)
    }{
    v_{f,j}^{(t)}(\tau)
    } .
    \label{eq:app_forward_reducible_fraction}
\end{equation}
The second is a forward-variance score that emphasizes policies with large reducible variance while downweighting uncertainty dominated by observation noise,
\begin{equation}
    s_{\mathrm{var}}^{(t)}(\tau)
    =
    \frac{1}{d_\phi}
    \sum_{j=1}^{d_\phi}
    \frac{
    \left(v_{\mathrm{epi},j}^{(t)}(\tau)\right)^2
    }{
    v_{f,j}^{(t)}(\tau)
    } .
    \label{eq:app_forward_reducible_score}
\end{equation}
These quantities are used in Boundary Lookahead to select target policies and
candidate-shortlist policies whose forward uncertainty can be reduced through execution.
The acquisition-specific one-step target-set update used for lookahead is described in
Appendix~\ref{app:active_querying}.

\paragraph{One-step forward update on a target set.}
Our active querying requires estimating how executing a candidate policy \(\tau\) would reduce forward uncertainty over a target policy \(\zeta\). The implementation uses the KRR analogue of a one-step Gaussian conditioning update~\cite{rasmussen2006gaussian}. Define the residual kernel cross term
\begin{equation}
    c_t(\zeta,\tau)
    =
    k_{\ell_f}(\zeta,\tau)
    -
    k_t(\zeta)^\top A_t^{-1}k_t(\tau).
    \label{eq:app_forward_cross}
\end{equation}
For outcome coordinate \(j\), the corresponding covariance proxy is
\begin{equation}
    \operatorname{cov}_{t,j}(\zeta,\tau)
    =
    c_t(\zeta,\tau)\gamma_{t,j}^2 .
    \label{eq:app_forward_cross_cov}
\end{equation}
Let
\begin{equation}
    d_{t,j}(\tau)
    =
    \gamma_{t,j}^2\left(e_t(\tau)+\lambda_f+\epsilon_f\right)
    +
    v_{\mathrm{obs},j}^{(t)}
    \label{eq:app_forward_update_denom}
\end{equation}
be the effective denominator for conditioning on the candidate observation. Given a hypothetical realized outcome \(\phi\) from executing \(\tau\), the target mean update is
\begin{equation}
    \mu_{f,j}^{(t+1)}(\zeta\mid\tau,\phi)
    =
    \mu_{f,j}^{(t)}(\zeta)
    +
    \frac{
    \operatorname{cov}_{t,j}(\zeta,\tau)
    }{
    d_{t,j}(\tau)
    }
    \left(
    \phi_j-\mu_{f,j}^{(t)}(\tau)
    \right),
    \label{eq:app_forward_mean_update}
\end{equation}
and the target epistemic variance update is
\begin{equation}
    v_{\mathrm{epi},j}^{(t+1)}(\zeta\mid\tau)
    =
    \left[
    v_{\mathrm{epi},j}^{(t)}(\zeta)
    -
    \frac{
    \operatorname{cov}_{t,j}(\zeta,\tau)^2
    }{
    d_{t,j}(\tau)
    }
    \right]_+ .
    \label{eq:app_forward_variance_update}
\end{equation}
The total post-query target variance is then
\begin{equation}
    v_{f,j}^{(t+1)}(\zeta\mid\tau)
    =
    v_{\mathrm{epi},j}^{(t+1)}(\zeta\mid\tau)
    +
    v_{\mathrm{obs},j}^{(t)} .
    \label{eq:app_forward_total_post_variance}
\end{equation}
The mean-observation approximation used in the reported Boundary Lookahead implementation sets the hypothetical candidate outcome to
\(\phi=\mu_f^{(t)}(\tau)\), so the residual term in Eq.~\eqref{eq:app_forward_mean_update} is zero and the forward lookahead reduces to a variance collapse over the target set.

\begin{table}[h]
\centering
\small
\setlength{\tabcolsep}{4pt}
\caption{Learner-side forward KRR settings for the stationary forward model. Recency weighting for nonstationary motor adaptation is described separately.}
\label{tab:forward_krr_hyperparameters}
\begin{tabularx}{\linewidth}{@{}l>{\raggedright\arraybackslash}X@{}}
\toprule
Quantity & Setting \\
\midrule
Forward model & Exact/full RBF kernel ridge regression \\
Kernel & \(k_{\ell_f}(\tau,\tau')=\exp(-\|\tau-\tau'\|_2^2/(2\ell_f^2))\) \\
Refitting schedule & Exact KRR refit after each new policy--outcome observation \\
Hyperparameter reselection & Performed at evaluation checkpoints using a validation split \\
Validation fraction & \(20\%\) of available forward observations \\
RBF length-scale candidates & \(\ell_f\in\{0.3,0.4,0.5,0.6,0.7\}\) \\
Ridge candidates & \(\lambda_f\in\{10^{-9},10^{-7},10^{-5},10^{-4},10^{-3},10^{-2},10^{-1}\}\) \\
Tuning cap & At most \(1024\) observations used for hyperparameter selection \\
Warm-up size & \(16\) observations before reliable hyperparameter tuning \\
Numerical jitter & \(\epsilon_f=10^{-10}\) \\
Variance floor & \(10^{-4}\) \\
Predictive sampling & Diagonal Gaussian samples clipped to \([0,1]^{d_\phi}\) \\
\bottomrule
\end{tabularx}
\end{table}

\subsection{Recency-weighted forward learning under outcome drift}
\label{app:recency_weighted_forward}

When human behavior adapts over repeated robot interactions, past policy--outcome observations may no longer be equally representative of the current response regime. The learner, however, does not observe the user's true adaptation progress or time constant. It only observes the temporally ordered forward history
\(\mathcal D_f^{(t)}=\{(\tau_i,\phi_i)\}_{i=1}^{t}\), together with the query index at which each observation was collected. SFIHILO therefore uses recency-weighted KRR to emphasize observations that are closer to the current response regime~\cite{klinkenberg2004learning}.

Let \(c_i\) denote the query index associated with observation \(i\). For a candidate decay time \(T_d\), we define an estimated adaptation-progress coordinate
\begin{equation}
    \hat\alpha_i(T_d)
    =
    1-\exp(-c_i/T_d),
    \qquad
    \hat\alpha_t(T_d)
    =
    1-\exp(-c_t/T_d),
    \label{eq:app_estimated_progress}
\end{equation}
where \(c_t\) is the current query index. Given a bandwidth \(h_d\), each observation receives the recency weight
\begin{equation}
    \tilde\omega_i^{(t)}
    =
    \max\!\left\{
    \exp\!\left(
    -\frac{
    \left(\hat\alpha_i(T_d)-\hat\alpha_t(T_d)\right)^2
    }{2h_d^2}
    \right),
    \omega_{\min}
    \right\},
    \qquad
    \omega_i^{(t)}
    =
    \frac{
    \tilde\omega_i^{(t)}
    }{
    t^{-1}\sum_{\ell=1}^{t}\tilde\omega_\ell^{(t)}
    } .
    \label{eq:app_recency_weights}
\end{equation}
The floor \(\omega_{\min}\) prevents older observations from being completely discarded, and the normalization keeps the average sample weight equal to one. This weighting controls the effective memory of the forward regressor: overly broad weights can lag behind the current response regime, whereas overly narrow weights can discard useful data and increase variance.

The weighted forward model is fit by replacing the stationary KRR objective with
\begin{equation}
    \hat f_t
    \in
    \arg\min_{f\in\mathcal H_{k_{\ell_f}}}
    \sum_{i=1}^{t}
    \omega_i^{(t)}
    \left\|
    \phi_i-f(\tau_i)
    \right\|_2^2
    +
    \lambda_f
    \|f\|_{\mathcal H_{k_{\ell_f}}}^2 .
    \label{eq:app_recency_weighted_krr}
\end{equation}
In the exact KRR implementation, this corresponds to replacing the unweighted kernel system with a weighted system using \(W_t=\operatorname{diag}(\omega_1^{(t)},\ldots,\omega_{t}^{(t)})\):
\begin{equation}
    A_t
    =
    K_t
    +
    (\lambda_f+\epsilon_f)W_t^{-1}.
    \label{eq:app_weighted_krr_system}
\end{equation}

Because the appropriate recency schedule is user-specific, the learned-decay variant does not fix \(T_d\) or \(h_d\) a priori. At hyperparameter-reselection checkpoints, it selects the decay schedule and KRR hyperparameters jointly by minimizing validation RMSE:
\begin{equation}
    (\hat T_d,\hat h_d,\hat\ell_f,\hat\lambda_f)
    \in
    \arg\min_{T_d,h_d,\ell_f,\lambda_f}
    \mathrm{RMSE}_{\mathrm{val}}
    \!\left(
    \mathrm{KRR}_{T_d,h_d,\ell_f,\lambda_f};
    \mathcal D_f^{(t)}
    \right).
    \label{eq:app_learned_decay_selection}
\end{equation}
The recency-weighted hyperparameter search was implemented using the
tree-structured Parzen estimator (TPE) procedure~\cite{bergstra2011algorithms}.
The unweighted baseline only reselects \((\ell_f,\lambda_f)\) and therefore uses
the exhaustive stationary KRR validation grid, whereas the recency-weighted
variants add continuous recency parameters and use TPE for the resulting
higher-dimensional search. All variants use the same validation-RMSE objective
and reselection checkpoints. Between reselection checkpoints, the selected
parameters are reused while the current progress coordinate advances with query count.

We compare three variants in the nonstationary experiments (Fig.~\ref{fig:final_results}G). The unweighted baseline sets \(\omega_i^{(t)}=1\), ignoring drift. The ground-truth-decay diagnostic uses the simulated user's true adaptation progress to define the recency weights; this variant is not available to SFIHILO, but serves as a reference for an ideal recency schedule while still reselecting the remaining forward-model hyperparameters. The learned-decay variant estimates the effective schedule from validation data using only the ordered query history. As shown in Fig.~\ref{fig:final_results}G, learned decay closely matched the ground-truth-decay performance curve while improving over the unweighted baseline, indicating that SFIHILO can recover an effective user-specific adaptation schedule without observing the true motor-adaptation time constant. Because the nonstationary experiment uses the default outcome-measurement noise
\((\sigma_\phi=0.05)\), late-stage RMSE is partially noise-limited: recency weighting
can reduce drift-induced bias, but it cannot remove finite-sample error induced by
noisy outcome observations.

\begin{table}[h]
\centering
\small
\setlength{\tabcolsep}{4pt}
\caption{Recency-weighted forward-learning settings for nonstationary experiments.}
\label{tab:recency_weighting_hyperparameters}
\begin{tabularx}{\linewidth}{@{}l>{\raggedright\arraybackslash}X@{}}
\toprule
Quantity & Setting \\
\midrule
Observed temporal variable & Query index \(c_i\) for each forward observation \\
Estimated progress coordinate & \(\hat\alpha_i(T_d)=1-\exp(-c_i/T_d)\) \\
Recency schedule parameters & \(T_d\) and \(h_d\) \\
Weight floor & \(\omega_{\min}=10^{-4}\) \\
Learned-decay selection & Joint validation-based selection of \(T_d,h_d,\ell_f,\lambda_f\) \\
Search method & TPE hyperparameter search~\cite{bergstra2011algorithms} \\
TPE trials & \(64\) trials with \(16\) startup trials \\
Decay-time search range & \(T_d\in[15,400]\) queries \\
Bandwidth search range & \(h_d\in[0.02,0.40]\) \\
Forward KRR parameters & \(\ell_f,\lambda_f\), using the same KRR search ranges as Table~\ref{tab:forward_krr_hyperparameters} \\
Compared variants & Unweighted, ground-truth decay, learned decay \\
\bottomrule
\end{tabularx}
\end{table}

\subsection{Inverse GP posterior computation}
\label{app:inverse_gp}

For the learner-side inverse reward model, we use an RBF-GP prior to impose a
smooth kernel-based inductive bias for modeling potential nonlinearities in the
outcome-space preference landscape. In query-limited human-in-the-loop settings,
this matched inductive bias can make nonlinear reward modeling more data-efficient
than using a high-capacity neural network: the kernel restricts the hypothesis space to
functions consistent with the expected smooth structure of human outcome preferences,
while maintaining tractable Bayesian uncertainty for active learning, as noted
in~\cite{biyik2024active}. The inverse model approximates the RBF-GP reward prior in
Eq.~(3) using random Fourier features (RFFs) and computes a tractable posterior over
the finite-dimensional parameters \(\Theta=(w,\eta)\), where \(w\) are the reward weights
and \(\eta\) is the log-choice sharpness.

For \(m\) random Fourier features, the feature map is
\begin{equation}
    z(\phi)
    =
    \sqrt{\frac{2}{m}}
    \cos\!\left(
    \frac{\Omega\phi}{\ell_r}+b
    \right),
    \qquad
    \Omega_{kj}\sim\mathcal N(0,1),
    \quad
    b_k\sim\operatorname{Unif}(0,2\pi),
    \label{eq:app_inverse_rff_features}
\end{equation}
with reward
\begin{equation}
    r_w(\phi)=w^\top z(\phi),
    \qquad
    w\sim\mathcal N(0,\sigma_w^2 I_m),
    \qquad
    \eta\sim\mathcal N(\mu_\eta,\sigma_\eta^2),
    \qquad
    \beta=e^\eta .
    \label{eq:app_inverse_reward_prior}
\end{equation}
Given the inverse-learning history
\(\mathcal D_r^{(t)}=\{(\bar\phi_i,\phi_i,I_i)\}_{i=1}^{t}\),
where \(I_i\in\{+1,-1\}\), define
\begin{equation}
    \delta z_i
    =
    z(\bar\phi_i)-z(\phi_i).
    \label{eq:app_inverse_feature_difference}
\end{equation}
The preference likelihood for comparison \(i\) is then
\begin{equation}
    p(I_i\mid \bar\phi_i,\phi_i,\Theta)
    =
    \sigma\!\left(
    I_i e^\eta \delta z_i^\top w
    \right),
    \label{eq:app_inverse_pairwise_likelihood}
\end{equation}
where \(\sigma(\cdot)\) is the logistic function.

The maximum a posteriori (MAP) estimate \(\hat\Theta=(\hat w,\hat\eta)\) minimizes the negative log posterior
\begin{equation}
    \mathcal L(w,\eta)
    =
    \sum_{i=1}^{t}
    \operatorname{softplus}\!\left(
    -I_i e^\eta \delta z_i^\top w
    \right)
    +
    \frac{1}{2\sigma_w^2}\|w\|_2^2
    +
    \frac{1}{2\sigma_\eta^2}(\eta-\mu_\eta)^2 .
    \label{eq:app_inverse_map_objective}
\end{equation}
We optimize Eq.~\eqref{eq:app_inverse_map_objective} using L-BFGS with warm starts across successive query updates~\cite{liu1989limited}.

Around the MAP solution, we construct a Gaussian approximation using a positive-semidefinite
local curvature matrix. Let
\begin{equation}
    u_i(\Theta)
    =
    I_i e^\eta \delta z_i^\top w,
    \qquad
    g_i
    =
    \nabla_\Theta u_i(\Theta)\big|_{\Theta=\hat\Theta},
    \qquad
    \lambda_i
    =
    \sigma(\hat u_i)\bigl(1-\sigma(\hat u_i)\bigr),
    \label{eq:app_inverse_curvature_terms}
\end{equation}
where \(\hat u_i=u_i(\hat\Theta)\). Stacking \(g_i^\top\) as rows of \(G\), the approximate posterior precision is
\begin{equation}
    \Lambda_t
    =
    \operatorname{diag}(\sigma_w^{-2}I_m,\sigma_\eta^{-2})
    +
    G^\top
    \operatorname{diag}(\lambda_i)
    G
    +
    \epsilon I .
    \label{eq:app_inverse_precision}
\end{equation}
The inverse posterior is approximated as
\begin{equation}
    p_r^{(t)}(\Theta)
    \approx
    \mathcal N\!\left(
    \hat\Theta,
    \Lambda_t^{-1}
    \right),
    \label{eq:app_inverse_laplace_posterior}
\end{equation}
with eigenvalue clipping applied for numerical stability before drawing posterior samples.

Given posterior samples
\(\Theta_s=(w_s,\eta_s)\), \(s=1,\ldots,S\), the posterior predictive preference probability used for evaluation and acquisition is
\begin{equation}
    q_t(+1\mid\phi_a,\phi_b)
    \approx
    \frac{1}{S}
    \sum_{s=1}^{S}
    \sigma\!\left(
    e^{\eta_s}
    \left[
    r_{w_s}(\phi_a)-r_{w_s}(\phi_b)
    \right]
    \right),
    \quad
    q_t(-1\mid\phi_a,\phi_b)
    =
    1-q_t(+1\mid\phi_a,\phi_b).
    \label{eq:app_inverse_posterior_predictive}
\end{equation}
These probabilities instantiate the inverse-model primitive used in the mutual-information baseline and in the proposed active querying.

\begin{table}[h]
\centering
\small
\caption{Learner-side inverse GP hyperparameters. These settings define the inverse reward model used by SFIHILO, not the simulated-user reward function used to generate preference labels.}
\label{tab:inverse_gp_hyperparameters}
\begin{tabular}{ll}
\toprule
Quantity & Setting \\
\midrule
Reward representation & RBF-GP approximated by RFFs \\
RFF features & \(m=256\) \\
Reward-kernel length scale & \(\ell_r=0.65\) \\
RFF frequencies & \(\Omega_{kj}\sim\mathcal N(0,1)\) \\
RFF phases & \(b_k\sim\operatorname{Unif}(0,2\pi)\) \\
Reward-weight prior & \(w\sim\mathcal N(0,I_m)\) \\
Log-sharpness prior & \(\eta\sim\mathcal N(0,1)\) \\
MAP optimizer & L-BFGS with warm starts~\cite{liu1989limited} \\
L-BFGS iterations & \(50\) \\
Posterior approximation & Gaussian approximation around MAP \\
Posterior samples & \(S=512\) \\
Numerical jitter & \(\epsilon=10^{-6}\) \\
Eigenvalue clipping & Applied for stable covariance sampling \\
\bottomrule
\end{tabular}
\end{table}

\subsection{Boundary Lookahead implementation}
\label{app:active_querying}

\begin{table}[t]
\centering
\small
\setlength{\tabcolsep}{4pt}
\caption{Implemented Boundary Lookahead querying at round \(t\).}
\label{tab:boundary_lookahead_algorithm}
\begin{tabularx}{\linewidth}{@{}l>{\raggedright\arraybackslash}X@{}}
\toprule
Stage & Computation \\
\midrule
Input &
Anchor outcome \(\bar\phi_t\), forward belief \(p_f^{(t)}(\phi\mid\tau)\), reward posterior samples \(\{\Theta_s\}_{s=1}^{S}\), and histories \(\mathcal D_t=(\mathcal D_f^{(t)},\mathcal D_r^{(t)})\). \\

Pool &
Sample a Monte Carlo policy pool
\(\mathcal P_t=\{\tau^{(m)}\}_{m=1}^{M_{\mathrm{pool}}}\subset\mathcal T\). \\

Cheap scoring &
For each \(\tau\in\mathcal P_t\), compute the forward mean \(\mu_f^{(t)}(\tau)\), the uncertainty decomposition \(v_{f}^{(t)}=v_{\mathrm{epi}}^{(t)}+v_{\mathrm{obs}}^{(t)}\), the reducible-uncertainty fraction \(\rho_f^{(t)}(\tau)\), and the mean-outcome reward-side mutual information \(a_{\mathrm{MI,mean}}^{(t)}(\tau)\). Define
\[
s_{\mathrm{rank}}^{(t)}(\tau)
=
a_{\mathrm{MI,mean}}^{(t)}(\tau)\rho_f^{(t)}(\tau).
\] \\

Target set &
Select the target boundary set
\[
\mathcal Z_t
=
\operatorname{TopK}_{K_{\mathrm{tar}}}
\{s_{\mathrm{rank}}^{(t)}(\tau):\tau\in\mathcal P_t\}.
\]
This finite set approximates the current anchor-relative preference-boundary region. \\

Shortlist &
Construct a deduplicated candidate shortlist
\[
\mathcal S_t
=
\mathcal S_{\mathrm{main}}
\cup
\mathcal S_{\mathrm{var}}
\cup
\mathcal S_{\mathrm{rand}},
\]
where \(\mathcal S_{\mathrm{main}}\) contains top-ranked policies under \(s_{\mathrm{rank}}^{(t)}\), \(\mathcal S_{\mathrm{var}}\) contains policies with high reducible forward-variance score \(s_{\mathrm{var}}^{(t)}\), and \(\mathcal S_{\mathrm{rand}}\) is a random guard set. \\

Current utility &
Estimate \(\mathcal U_t(\mathcal Z_t)\) using reward posterior samples and \(M_{\mathrm{tar}}\) forward samples
\(\Phi_\zeta^{(m)}\sim p_f^{(t)}(\cdot\mid\zeta)\) for each \(\zeta\in\mathcal Z_t\). \\

Candidate approximation &
For each \(\tau\in\mathcal S_t\), use the mean-observation approximation
\[
\tilde\phi_\tau=\mu_f^{(t)}(\tau),
\]
collapsing the candidate-side forward integral while retaining target-set forward uncertainty. \\

Hypothetical inverse update &
Compute candidate label probabilities \(p_{s,\tau}^{+}=p_s^+(\bar\phi_t,\tilde\phi_\tau)\) and approximate the two label-conditioned inverse posteriors by importance weights
\[
\omega_s^{(+)}(\tau)
=
\frac{p_{s,\tau}^{+}}{\sum_{j=1}^{S}p_{j,\tau}^{+}},
\qquad
\omega_s^{(-)}(\tau)
=
\frac{1-p_{s,\tau}^{+}}{\sum_{j=1}^{S}(1-p_{j,\tau}^{+})}.
\] \\

Hypothetical forward update &
Apply the one-step KRR conditioning update to the target set. Under the mean-observation approximation, target means remain unchanged and target forward variances are reduced through the variance-collapse update. \\

Post-query utility &
Estimate \(\mathcal U_{t+1}^{(+)}(\mathcal Z_t\mid\tau)\) and \(\mathcal U_{t+1}^{(-)}(\mathcal Z_t\mid\tau)\) using the label-conditioned inverse weights and the updated target forward variances. \\

Selection &
Select
\[
\tau_t
\in
\arg\max_{\tau\in\mathcal S_t}
\left[
\mathcal U_t(\mathcal Z_t)
-
\bar p_\tau
\mathcal U_{t+1}^{(+)}(\mathcal Z_t\mid\tau)
-
(1-\bar p_\tau)
\mathcal U_{t+1}^{(-)}(\mathcal Z_t\mid\tau)
\right],
\]
where \(\bar p_\tau=S^{-1}\sum_{s=1}^{S}p_{s,\tau}^{+}\). \\
\bottomrule
\end{tabularx}
\end{table}

While the Mutual Information baseline evaluates reward-side informativeness at the current forward mean~\cite{houlsby2011bayesian}, Boundary Lookahead querying values an executable policy by its expected effect on preference-boundary resolution after updating both the forward and inverse beliefs. Here the reported implementation of Boundary Lookahead querying is described in detail.

\paragraph{Acquisition-side preference probabilities.}
Let \(\Theta_s=(w_s,\eta_s)\), \(s=1,\ldots,S\), denote posterior samples from the inverse reward model, with \(\beta_s=\exp(\eta_s)\). For an ordered outcome pair \((\phi_a,\phi_b)\), define the sample-specific reward gap
\begin{equation}
    \Delta_s(\phi_a,\phi_b)
    =
    r_{w_s}(\phi_a)-r_{w_s}(\phi_b).
    \label{eq:app_active_reward_gap}
\end{equation}
For acquisition scoring, we use the sample-specific probability of preferring \(\phi_a\) over \(\phi_b\)~\cite{bishop2006pattern,mackay1992evidence},
\begin{equation}
    p_s^+(\phi_a,\phi_b)
    =
    \sigma\!\left(
    \frac{
    \beta_s \Delta_s(\phi_a,\phi_b)
    }{
    \sqrt{1+\frac{\pi}{8}(\beta_s\sigma_\Delta)^2}
    }
    \right),
    \label{eq:app_active_noisy_preference_probability}
\end{equation}
where \(\sigma_\Delta=\sqrt{2}\sigma_{\tilde r}\) is the reward-gap noise standard deviation induced by independent reward-evaluation noise. In noise-free settings, \(\sigma_\Delta=0\), and this reduces to the standard logistic preference probability.

\paragraph{Candidate pool and cheap ranking scores.}
At each query round, the implementation samples a Monte Carlo policy pool
\begin{equation}
    \mathcal P_t
    =
    \{\tau^{(m)}\}_{m=1}^{M_{\mathrm{pool}}}
    \subset \mathcal T .
    \label{eq:app_active_pool}
\end{equation}
For each candidate \(\tau\in\mathcal P_t\), the forward model returns the predictive mean \(\mu_f^{(t)}(\tau)\) and the uncertainty decomposition \(v_{f,j}^{(t)}(\tau)=v_{\mathrm{epi},j}^{(t)}(\tau)+v_{\mathrm{obs},j}^{(t)}(\tau)\). The reward-side mean-outcome mutual information score is
\begin{equation}
    a_{\mathrm{MI,mean}}^{(t)}(\tau)
    =
    H\!\left(\bar p_t(\tau)\right)
    -
    \frac{1}{S}
    \sum_{s=1}^{S}
    H\!\left(
    p_s^+(\bar\phi_t,\mu_f^{(t)}(\tau))
    \right),
    \label{eq:app_active_mean_mi}
\end{equation}
where
\begin{equation}
    \bar p_t(\tau)
    =
    \frac{1}{S}
    \sum_{s=1}^{S}
    p_s^+(\bar\phi_t,\mu_f^{(t)}(\tau)).
    \label{eq:app_active_mean_pbar}
\end{equation}
To prioritize policies whose forward uncertainty can be reduced through execution, we combine this score with the reducible-uncertainty fraction defined in Appendix~\ref{app:forward_krr}:
\begin{equation}
    s_{\mathrm{rank}}^{(t)}(\tau)
    =
    a_{\mathrm{MI,mean}}^{(t)}(\tau)\,
    \rho_f^{(t)}(\tau).
    \label{eq:app_active_rank_score}
\end{equation}

\paragraph{Target set and candidate shortlist.}
The target set is the top-\(K_{\mathrm{tar}}\) subset of the policy pool under the ranking score:
\begin{equation}
    \mathcal Z_t
    =
    \operatorname{TopK}_{K_{\mathrm{tar}}}
    \left\{
    s_{\mathrm{rank}}^{(t)}(\tau):
    \tau\in\mathcal P_t
    \right\}.
    \label{eq:app_active_target_set}
\end{equation}
This set is a finite approximation to the current anchor-relative preference-boundary region. The one-step lookahead is not evaluated on the full pool. Instead, the implementation forms a deduplicated shortlist
\begin{equation}
    \mathcal S_t
    =
    \mathcal S_{\mathrm{main}}
    \cup
    \mathcal S_{\mathrm{var}}
    \cup
    \mathcal S_{\mathrm{rand}},
    \label{eq:app_active_shortlist}
\end{equation}
where \(\mathcal S_{\mathrm{main}}\) contains top-ranked candidates under \(s_{\mathrm{rank}}^{(t)}\), \(\mathcal S_{\mathrm{var}}\) contains candidates with high reducible forward-variance score \(s_{\mathrm{var}}^{(t)}\), and \(\mathcal S_{\mathrm{rand}}\) is a random guard set.

\paragraph{Current target-region utility.}
For each target policy \(\zeta\in\mathcal Z_t\), we sample target outcomes from the current forward belief,
\begin{equation}
    \Phi_\zeta^{(m)}
    \sim
    p_f^{(t)}(\cdot\mid\zeta),
    \qquad
    m=1,\ldots,M_{\mathrm{tar}}.
    \label{eq:app_active_target_samples}
\end{equation}
For each target forward sample and reward posterior sample, define
\begin{equation}
    p_{s,\zeta}^{(m)}
    =
    p_s^+(\bar\phi_t,\Phi_\zeta^{(m)}).
    \label{eq:app_active_target_probabilities}
\end{equation}
The implemented target-wise utility is
\begin{equation}
    \mathcal U_t(\zeta)
    =
    H\!\left(
    \frac{1}{M_{\mathrm{tar}}S}
    \sum_{m=1}^{M_{\mathrm{tar}}}
    \sum_{s=1}^{S}
    p_{s,\zeta}^{(m)}
    \right)
    -
    \frac{1}{M_{\mathrm{tar}}S}
    \sum_{m=1}^{M_{\mathrm{tar}}}
    \sum_{s=1}^{S}
    H\!\left(
    p_{s,\zeta}^{(m)}
    \right),
    \label{eq:app_active_targetwise_utility}
\end{equation}
and the target-region utility is
\begin{equation}
    \mathcal U_t(\mathcal Z_t)
    =
    \frac{1}{|\mathcal Z_t|}
    \sum_{\zeta\in\mathcal Z_t}
    \mathcal U_t(\zeta).
    \label{eq:app_active_current_utility}
\end{equation}
This estimates mutual information between the target-comparison label and the joint uncertainty induced by reward posterior samples and target forward-outcome samples.

\paragraph{Mean-observation candidate approximation.}
For a candidate query policy \(\tau\in\mathcal S_t\), the exact one-step lookahead would integrate over the unknown candidate outcome.  For tractability and to avoid compounding Monte Carlo noise in the outer candidate-outcome integral, the reported implementation uses the mean-observation approximation
\begin{equation}
    \tilde\phi_\tau
    =
    \mu_f^{(t)}(\tau).
    \label{eq:app_active_mean_observation}
\end{equation}
This approximation collapses only the candidate-outcome integral,
while retaining preference-label averaging and target-outcome
uncertainty (Eqs.~(62), (69), and (72)). Setting
$\tilde{\phi}_{\tau}=\mu_f^{(t)}(\tau)$ leaves target means unchanged
(Eq.~(37)) but retains covariance-dependent target variance updates
(Eq.~(38)). Thus, the approximation retains the central rationale
of Boundary Lookahead: valuing query-induced changes in joint
forward--inverse uncertainty over $Z_t$, whereas the Mutual Information baseline
evaluates current reward-side informativeness at the predicted
candidate outcome.

Given \(\tilde\phi_\tau\), define the candidate-side label probabilities
\begin{equation}
    p_{s,\tau}^{+}
    =
    p_s^+(\bar\phi_t,\tilde\phi_\tau),
    \qquad
    \bar p_\tau
    =
    \frac{1}{S}
    \sum_{s=1}^{S}
    p_{s,\tau}^{+}.
    \label{eq:app_active_candidate_label_probability}
\end{equation}
Rather than refitting the inverse model for each hypothetical label, the implementation performs a sample-based importance update~\cite{smith1992bayesian}:
\begin{equation}
    \omega_s^{(+)}(\tau)
    =
    \frac{
    p_{s,\tau}^{+}
    }{
    \sum_{j=1}^{S} p_{j,\tau}^{+}
    },
    \qquad
    \omega_s^{(-)}(\tau)
    =
    \frac{
    1-p_{s,\tau}^{+}
    }{
    \sum_{j=1}^{S} (1-p_{j,\tau}^{+})
    } .
    \label{eq:app_active_importance_weights}
\end{equation}
These weights approximate the inverse posterior after observing the two possible labels \(y\in\{+1,-1\}\).

\paragraph{Post-query target utility.}
The forward-side hypothetical update applies the one-step KRR conditioning formulas from Appendix~\ref{app:forward_krr} to the target set. Under the mean-observation approximation, the candidate residual is zero, so the target forward means remain unchanged while the target variances are reduced. We then draw post-query target samples
\begin{equation}
    \Phi_{\zeta,\tau}^{\mathrm{post},(m)}
    \sim
    \mathcal N\!\left(
    \mu_f^{(t)}(\zeta),
    \operatorname{diag}
    \left(
    v_f^{(t+1)}(\zeta\mid\tau)
    \right)
    \right),
    \qquad
    m=1,\ldots,M_{\mathrm{tar}}.
    \label{eq:app_active_post_target_samples}
\end{equation}
For a hypothetical label \(y\in\{+1,-1\}\), the post-query target-wise utility is
\begin{equation}
\begin{aligned}
    \mathcal U_{t+1}^{(y)}(\zeta\mid\tau)
    &=
    H\!\left(
    \frac{1}{M_{\mathrm{tar}}}
    \sum_{m=1}^{M_{\mathrm{tar}}}
    \sum_{s=1}^{S}
    \omega_s^{(y)}(\tau)
    p_s^+(\bar\phi_t,\Phi_{\zeta,\tau}^{\mathrm{post},(m)})
    \right)
    \\
    &\quad
    -
    \frac{1}{M_{\mathrm{tar}}}
    \sum_{m=1}^{M_{\mathrm{tar}}}
    \sum_{s=1}^{S}
    \omega_s^{(y)}(\tau)
    H\!\left(
    p_s^+(\bar\phi_t,\Phi_{\zeta,\tau}^{\mathrm{post},(m)})
    \right).
\end{aligned}
\label{eq:app_active_post_targetwise_utility}
\end{equation}
The regional post-query utility is
\begin{equation}
    \mathcal U_{t+1}^{(y)}(\mathcal Z_t\mid\tau)
    =
    \frac{1}{|\mathcal Z_t|}
    \sum_{\zeta\in\mathcal Z_t}
    \mathcal U_{t+1}^{(y)}(\zeta\mid\tau).
    \label{eq:app_active_post_region_utility}
\end{equation}

\paragraph{Implemented Boundary Lookahead score.}
The final implemented acquisition score is
\begin{equation}
    a_{\mathrm{BL}}^{(t)}(\tau)
    =
    \mathcal U_t(\mathcal Z_t)
    -
    \bar p_\tau
    \mathcal U_{t+1}^{(+)}(\mathcal Z_t\mid\tau)
    -
    (1-\bar p_\tau)
    \mathcal U_{t+1}^{(-)}(\mathcal Z_t\mid\tau),
    \label{eq:app_active_meanobs_score}
\end{equation}
and the selected policy is
\begin{equation}
    \tau_t
    \in
    \arg\max_{\tau\in\mathcal S_t}
    a_{\mathrm{BL}}^{(t)}(\tau).
    \label{eq:app_active_selected_query}
\end{equation}
Thus, the implementation keeps mutual information as the inverse-learning primitive, but changes the object being valued: instead of asking how informative one isolated comparison would be, it asks how much executing a policy is expected to reduce uncertainty over the unresolved anchor-relative boundary region.

In our single-threaded CPU implementation, Boundary Lookahead required 1.15 s per query, compared with 0.78 s for the Mutual Information baseline, allowing query selection between physical trials.

\FloatBarrier
\begin{table}[h]
\centering
\small
\setlength{\tabcolsep}{4pt}
\caption{Boundary Lookahead implementation settings.}
\label{tab:boundary_lookahead_hyperparameters}
\begin{tabularx}{\linewidth}{@{}l>{\raggedright\arraybackslash}X@{}}
\toprule
Quantity & Setting \\
\midrule
Candidate pool size & \(M_{\mathrm{pool}}=\max(4000,400d_\tau)\) \\
Target set size & \(K_{\mathrm{tar}}=32\) \\
Target forward samples & \(M_{\mathrm{tar}}=3\) \\
Candidate outcome approximation & \(\tilde\phi_\tau=\mu_f^{(t)}(\tau)\) \\
Candidate-side forward integral & Collapsed to one mean observation \\
Target-set forward uncertainty & Retained through forward sampling and one-step variance collapse \\
Main shortlist size & \(K_{\mathrm{main}}=16\) \\
Forward-variance shortlist size & \(K_{\mathrm{var}}=4\) \\
Random guard shortlist size & \(K_{\mathrm{rand}}=4\) \\
Continuous soft proposals & Not used in the reported mean-observation results \\
Inverse hypothetical update & Label-conditioned importance reweighting of reward posterior samples \\
Forward hypothetical update & One-step KRR variance collapse over the target set \\
Selected query & \(\tau_t\in\arg\max_{\tau\in\mathcal S_t} a_{\mathrm{BL}}^{(t)}(\tau)\) \\
\bottomrule
\end{tabularx}
\end{table}
\FloatBarrier

\subsection{Final policy reward-gap evaluation}
\label{app:policy-reward-gap}

\paragraph{Pool-based final policy reward-gap diagnostic.}

We use a fixed evaluation pool for this diagnostic because a continuous-space reward gap would require solving two nonconvex optimization problems with comparable global accuracy: the oracle reference optimization
\[
\max_{\tau\in\mathcal T} r^\star(f_{\star,T}(\tau))
\]
and the learned-model final policy problem in Eq. (\ref{eq:final_policy_optimization}) for comparison. Any optimization error in either downstream problem would be conflated with the quality of the learned forward--inverse model, making the comparison depend on optimizer initialization or budget rather than on the active querying rule. The common evaluation pool removes this nuisance variable: all querying methods rank the same feasible policies and are compared against the same clean-reward pool optimum.

Thus, let
\[
\mathcal T_{\mathrm{eval}}
=
\{\tau^{(1)},\ldots,\tau^{(N_{\mathrm{eval}})}\}
\subset \mathcal T
\]
denote a fixed evaluation pool of feasible policies. Throughout this diagnostic, \(T\) denotes the terminal evaluation checkpoint, \(T=q_{\max}=1000\). For each policy in this pool, we compute the clean oracle reward
\[
R^\star_T(\tau)
=
r^\star(f_{\star,T}(\tau)).
\]
The pool optimum is therefore
\[
\tau^\star_{\mathrm{eval}}
\in
\arg\max_{\tau\in\mathcal T_{\mathrm{eval}}}
R^\star_T(\tau).
\]

After \(T\) active queries, the learned reward posterior is summarized by
\[
\bar r_T(\phi)
=
\mathbb E_{\Theta\sim p_r^{(T)}}[r_w(\phi)],
\]
as in the final policy-selection rule in Eq.~(\ref{eq:final_policy_optimization}). To match the forward-belief expectation in Eq.~(\ref{eq:final_policy_optimization}), we use the Monte Carlo policy-selection score
\[
\widehat V_T^{\mathrm{MC}}(\tau)
=
\frac{1}{M_{\mathrm{MC}}}
\sum_{m=1}^{M_{\mathrm{MC}}}
\bar r_T(\phi_{\tau}^{(m)}),
\qquad
\phi_{\tau}^{(m)}
\sim
p_f^{(T)}(\cdot\mid \tau).
\]
The selected policy is
\[
\hat\tau_T^{\mathrm{MC}}
\in
\arg\max_{\tau\in\mathcal T_{\mathrm{eval}}}
\widehat V_T^{\mathrm{MC}}(\tau),
\]
and the reported final policy reward gap is
\[
G_T^{\mathrm{MC}}
=
R^\star_T(\tau^\star_{\mathrm{eval}})
-
R^\star_T(\hat\tau_T^{\mathrm{MC}}).
\]
Lower values indicate that the learned forward--inverse model selected a policy whose clean oracle reward is closer to the best policy in the evaluation pool. In our experiments, the same \(N_{\mathrm{eval}}=50{,}000\) evaluation policies were used for all querying methods within each run, and \(M_{\mathrm{MC}}=16\) forward samples per policy were used for the Monte Carlo policy-selection score.

\paragraph{Policy reoptimization across feasible sets.}
We constructed \(210=\binom{8}{4}\times3\) feasible sets. For each four-parameter subset, we selected a reward-independent assignment of the low, middle, and high thirds of \([0,1]\) and generated two complementary assignments by rotating every range label as low \(\rightarrow\) middle \(\rightarrow\) high \(\rightarrow\) low. Thus, each constrained parameter occupied each third exactly once across the three assignments, while the remaining four parameters were unrestricted. At the 1,000-query checkpoint, without collecting additional queries or updating either model, we applied the above policy-selection rule within each feasible subset \(\mathcal F\subset\mathcal T_{\mathrm{eval}}\). Performance was measured by the clean reward gap between the selected policy and the best policy in the same subset. We averaged this gap across all 210 sets for each simulated user and then compared methods across the 30 users, treating users rather than feasible sets as the independent statistical units.
\end{document}